\ifdefined\XeTeXrevision\else
\pdfoutput=1
\fi
\documentclass[11pt]{article}

\usepackage[utf8]{inputenc}
\usepackage[T1]{fontenc}
\usepackage{microtype}
\usepackage{amsmath,amssymb,amsfonts}
\usepackage{bm}
\usepackage{graphicx}
\usepackage{booktabs}
\usepackage{multirow}
\usepackage{natbib}
\usepackage{hyperref}
\usepackage{cleveref}
\usepackage{xcolor}
\usepackage{enumitem}
\usepackage{tabularx}
\usepackage{caption}
\usepackage{subcaption}
\usepackage{algorithm}
\usepackage{algpseudocode}
\usepackage[margin=1in]{geometry}
\usepackage{url}
\usepackage{makecell}
\usepackage{pifont}
\newcommand{\cmark}{\ding{51}}
\newcommand{\xmark}{\ding{55}}

\newcommand{\benchmark}{\textsc{FinVerBench}}
\hypersetup{
  colorlinks=true,
  linkcolor=blue,
  citecolor=blue,
  urlcolor=blue
}

\title{\benchmark{}: Benchmark Validity and Calibration \\in Large Language Model Financial Statement Verification}

\newif\ifanonymized
\anonymizedfalse

\ifanonymized
\author{Anonymous Author(s)}
\else
\author{Silu Panda\\\texttt{pandasilu428@gmail.com}}
\fi

\date{}

\begin{document}
\maketitle

\begin{abstract}
We introduce \benchmark{}, a benchmark and validity study for financial statement verification: determining whether a set of corporate financial statements is numerically consistent from the information shown to the model.
\benchmark{} is built from SEC 10-K XBRL filings for 43 S\&P 500 companies and defines a four-category error taxonomy covering arithmetic, cross-statement linkage, year-over-year, and magnitude perturbations.
We attempt fifteen contemporary LLM evaluations and report fourteen complete runs; a Gemini 2.5 Pro run is excluded from the main comparison because 40/108 gateway calls failed.
All binary metrics exclude underdetermined positive instances whose perturbed line item is not rendered, leaving a 105-instance observable diagnostic subset (43 clean, 62 error-injected).
Under the original guided-checklist prompt on the unrounded diagnostic subset, nine of fourteen complete LLM runs produce 95--100\% false positives on clean statements, while one run achieves 0\% observed false positives.
Benchmark rendering choices materially affect measured recall: on a realistic rounded variant of the same observable subset, the calibrated model's recall is 79.0\% with 0\% observed FPR, compared with 100.0\% recall on the unrounded diagnostic variant.
These results support a construct-validity conclusion rather than a final leaderboard: financial statement verification is not merely arithmetic detection, but calibrated judgment under incomplete observability, prompt-induced assumptions, and realistic numerical rendering.
\ifanonymized Code and data will be released upon acceptance.\else \benchmark{} and all code are publicly available.\fi
\end{abstract}

\section{Introduction}
\label{sec:intro}

Financial statements are the primary mechanism through which corporations communicate their economic position to investors, regulators, and the public.
Their usefulness depends on internal consistency: assets must equal liabilities plus equity, net income must reconcile across statements, the cash flow statement must tie to changes in the balance sheet, and numerous other accounting relationships must hold simultaneously.
In practice, these relationships are checked through internal controls, external audits, and regulatory review~\citep{auditllm2025}.

Large language models (LLMs) are now used to summarize, search, and analyze financial filings~\citep{wu2023bloomberggpt, lin2026openfinllm}.
This adoption raises a natural question: \emph{can LLMs reliably verify whether a set of financial statements is numerically consistent?}
The question is practically important for two reasons.
First, a model that cannot detect basic inconsistencies may propagate or amplify errors in downstream analyses---a risk that \citet{failsafeqa2025} show persists even for strong long-context financial QA systems.
Second, the scale of public filings exceeds what human reviewers can exhaustively inspect, while XBRL tagging errors remain common in SEC data~\citep{fintagging2025}.
Reliable automated verification would therefore be valuable both as a safety layer for LLM-based financial tooling and as an auditing aid.

Existing evaluation benchmarks, while valuable, do not directly isolate this question.
Benchmarks such as FinQA~\citep{chen2021finqa}, TAT-QA~\citep{zhu2021tatqa}, and FinanceBench~\citep{islam2023financebench} evaluate LLMs as \emph{generators}---producing answers to questions about financial data.
More recent benchmarks like FinBen~\citep{xie2024finben}, DocFinQA~\citep{reddy2024docfinqa}, and FinanceReasoning~\citep{tang2025financereasoning} extend this paradigm to more complex reasoning tasks but remain fundamentally question-answering evaluations.
Even benchmarks that touch on auditing, such as FinAuditing~\citep{wang2025finauditing}, primarily evaluate taxonomy-grounded structured reasoning over XBRL filings rather than controlled perturbation-based consistency verification across rendered financial statements.

We treat \emph{verification} and \emph{generation} as distinct tasks that warrant separate evaluation.
Generation requires producing a correct answer from context; verification requires assessing whether a given set of numerical relationships is internally consistent.
The latter demands systematic checking of multiple interdependent constraints---a capability that is not isolated by existing financial QA benchmarks but is essential for trustworthy financial AI.

\paragraph{Contributions.}
We make the following contributions:
\begin{enumerate}[topsep=2pt,itemsep=1pt]
    \item We formalize \textbf{financial statement verification} as a distinct evaluation task for LLMs, centered on detecting, localizing, and explaining numerical inconsistencies in structured financial documents (\Cref{sec:task}).
    \item We define a \textbf{four-category, twelve-subtype error taxonomy} grounded in auditing practice and instantiate eleven subtypes in the current benchmark; the CL-D\&A subtype is specified but excluded from the main evaluation because XBRL depreciation reporting is insufficiently consistent (\Cref{sec:taxonomy}).
    \item We present \benchmark{}, a \textbf{generated benchmark corpus} of 1,985 instances constructed from real SEC 10-K filings through controlled error injection at six magnitude levels, together with an observable binary subset that relabels hidden-field perturbations as not-enough-information (\Cref{sec:benchmark}).
    \item We evaluate a \textbf{deterministic rule-based verifier} on both the full corpus and the shared LLM diagnostic subset, and report fourteen complete \textbf{frontier LLM} runs plus one incomplete Gemini run, showing that model performance under the original guided checklist is governed primarily by calibration (\Cref{sec:results}).
    \item We identify a \textbf{verification noise floor} induced by partial observability in XBRL-based financial data, and we show through a \textbf{rounding analysis} that rendering choices materially affect measured verification performance (\Cref{sec:analysis}).
\end{enumerate}

\section{Related Work}
\label{sec:related}

\subsection{Financial NLP Benchmarks}
\label{sec:related:benchmarks}

The landscape of financial NLP benchmarks has expanded rapidly since 2021.
FinQA~\citep{chen2021finqa} introduced 8,281 question-answer pairs over 2,789 earnings reports, with annotated reasoning programs showing that pre-trained models fell far short of expert human performance on multi-step numerical reasoning.
TAT-QA~\citep{zhu2021tatqa} extended this to hybrid tabular-textual content, and ConvFinQA~\citep{chen2022convfinqa} added a conversational dimension.

FinanceBench~\citep{islam2023financebench} scaled to 10,231 questions about publicly traded companies and revealed that GPT-4-Turbo with retrieval incorrectly answered or refused 81\% of questions---a finding that underscored the gap between general-purpose language capabilities and domain-specific financial reasoning.

The most comprehensive effort to date is FinBen~\citep{xie2024finben}, which aggregates 42 datasets across 24 financial NLP tasks organized into seven categories.
FinBen demonstrated that while LLMs excel at information extraction and textual analysis, they struggle with advanced reasoning, forecasting, and text generation tasks.
FinanceReasoning~\citep{tang2025financereasoning} further isolated numerical reasoning, finding that even OpenAI o1 with Proof-of-Thought achieves only 89.1\% accuracy, with performance degrading significantly on multi-formula problems.
DocMath-Eval~\citep{zhao2024docmatheval} assessed 48 LLMs on numerical reasoning over long documents, finding that even GPT-4o significantly lags behind human experts.
FinanceMATH~\citep{zhao2024financemath} focused on knowledge-intensive math problems, where the best system achieved only 60.9\% accuracy versus an estimated 92\% human expert performance.

More recently, specialized benchmarks have targeted specific gaps: DocFinQA~\citep{reddy2024docfinqa} extended financial QA to long contexts (averaging 123K words); SECQUE~\citep{benyoash2025secque} focused on SEC filing analysis with expert-written questions; Fin-RATE~\citep{jiang2026finrate} introduced longitudinal and cross-entity comparisons, revealing 18.6\% accuracy drops on temporal analysis; and FinSheet-Bench~\citep{finsheetbench2026} found that even the best model (Gemini 2.5 Pro) degrades from 82.4\% to 48.6\% accuracy on complex financial spreadsheets, suggesting fundamental architectural limitations.

Despite this proliferation, most existing financial benchmarks still evaluate LLMs as \emph{answer generators}.
The closest work to verification is FinDVer~\citep{wei2024findver}, which provides 2,400 expert-annotated examples for claim verification over financial documents, and FISCAL~\citep{nair2025fiscal}, which trains lightweight verifiers on synthetic claim-document pairs.
However, both frame verification as a classification problem (supported/refuted/not-enough-info) rather than as numerical consistency checking across multiple interrelated financial statements.
To the best of our knowledge, prior financial benchmarks do not directly evaluate whether a presented bundle of financial statements is internally consistent under controlled numeric perturbations.

\subsection{LLMs for Financial Analysis and Auditing}
\label{sec:related:analysis}

Domain-specific financial LLMs, including BloombergGPT~\citep{wu2023bloomberggpt} (50B parameters, trained on 363B tokens of financial data) and FinGPT~\citep{yang2023fingpt} (an open-source alternative), have demonstrated competence on standard financial NLP tasks, though the extent to which this reflects genuine reasoning versus memorization remains debated~\citep{memorization2025}.

In the auditing domain, \citet{auditllm2025} proposed a five-stage evaluation framework and found that LLMs can identify errors in financial statements but struggle to explain them or cite relevant accounting standards.
FinAuditing~\citep{wang2025finauditing} introduced a taxonomy-structured multi-document benchmark from real XBRL filings, testing semantic matching, relationship extraction, and mathematical reasoning, with accuracy dropping by 60--90\% on hierarchical multi-document structures.
It therefore overlaps with \benchmark{} in its use of XBRL and taxonomy-grounded numerical reasoning.
\benchmark{} differs by focusing specifically on controlled perturbation-based consistency verification across rendered financial statements, with explicit error magnitudes and clean-instance calibration analysis.
\citet{compliance2025} evaluated LLMs for regulatory compliance verification, finding that Llama-2 70B showed strong non-compliance detection.

Most closely related to our work is AuditBench~\citep{wang2025auditbench}, which also injects errors into financial statements and benchmarks LLM detection capabilities.
However, AuditBench differs in three key aspects: it uses image-based tables converted to text rather than native XBRL filings; it focuses on transaction-level auditing with journal entries rather than cross-statement consistency; and it does not systematically vary error magnitude to characterize detection sensitivity curves.
\benchmark{} complements AuditBench by providing XBRL-native, cross-statement verification at controlled magnitudes.

\subsection{LLM Hallucination and Numerical Reliability}
\label{sec:related:hallucination}

The reliability of LLMs on numerical tasks in finance has been called into question on multiple fronts.
\citet{kang2023hallucination} found serious hallucination in financial tasks, and FAITH~\citep{zhang2025faith} showed that top models collapse from 95.6\% accuracy on simple lookups to near 0\% on multivariate calculations.
\citet{memorization2025} demonstrated that perturbing financial statements drops LLM predictive accuracy to random chance, suggesting memorization rather than genuine reasoning.
The ``arithmetic gap'' in financial LLMs~\citep{arithmeticgap2026} and the finding that LLMs cannot reliably detect their own reasoning errors~\citep{mathcomp2025} further underscore the fragility of numerical computation.
\citet{venra2026} propose neuro-symbolic approaches with deterministic fact ledgers, arguing that LLMs must be ``surgically'' relieved of arithmetic responsibility.

These findings motivate our benchmark: if LLMs hallucinate and make arithmetic errors in generation tasks, how effectively can they \emph{detect} errors that are already present in financial documents?

\subsection{Positioning of \benchmark{}}
\label{sec:related:positioning}

\benchmark{} differs from prior work along three key dimensions:
\begin{enumerate}[topsep=2pt,itemsep=1pt]
    \item \textbf{Task}: Verification (error detection in given data) rather than generation (answer production).
    \item \textbf{Error control}: Errors are systematically injected at known types and magnitudes, enabling fine-grained sensitivity analysis.
    \item \textbf{Cross-statement scope}: The benchmark tests inter-statement consistency (e.g., income statement $\leftrightarrow$ balance sheet linkages) under controlled perturbations and explicit clean-instance calibration analysis.
\end{enumerate}

\Cref{tab:benchmark_comparison} summarizes how \benchmark{} relates to existing benchmarks.

\begin{table}[t]
\centering
\small
\caption{Comparison of \benchmark{} with existing financial NLP benchmarks. To the best of our knowledge, \benchmark{} is the first benchmark to jointly combine XBRL-derived real filings, controlled numeric perturbations, cross-statement consistency checks, and magnitude-sensitive calibration analysis.}
\label{tab:benchmark_comparison}
\begin{tabular}{lccccc}
\toprule
\textbf{Benchmark} & \textbf{Task} & \textbf{Cross-Stmt} & \textbf{Error Ctrl} & \textbf{Mag. Vary} & \textbf{Source} \\
\midrule
FinQA~\citeyearpar{chen2021finqa} & QA & \xmark & \xmark & \xmark & Earnings Reports \\
FinanceBench~\citeyearpar{islam2023financebench} & QA & \xmark & \xmark & \xmark & SEC Filings \\
FinBen~\citeyearpar{xie2024finben} & Multi-task & \xmark & \xmark & \xmark & Multiple \\
DocMath-Eval~\citeyearpar{zhao2024docmatheval} & QA & \xmark & \xmark & \xmark & Long Docs \\
FinDVer~\citeyearpar{wei2024findver} & Claim Ver. & \xmark & \xmark & \xmark & Annual Reports \\
FinAuditing~\citeyearpar{wang2025finauditing} & IE/MR & Partial & \xmark & \xmark & XBRL Filings \\
FAITH~\citeyearpar{zhang2025faith} & Halluc. & \xmark & \xmark & \xmark & Annual Reports \\
AuditBench~\citeyearpar{wang2025auditbench} & Audit & \xmark & \cmark & \xmark & Tables + Txns \\
FinSheet-Bench~\citeyearpar{finsheetbench2026} & Spreadsheet & \xmark & \xmark & \xmark & Synthetic \\
\midrule
\benchmark{} & \textbf{Verif.} & \cmark & \cmark & \cmark & \textbf{10-K XBRL} \\
\bottomrule
\end{tabular}
\end{table}

\section{\benchmark{} Benchmark Design}
\label{sec:benchmark}

\subsection{Task Formulation}
\label{sec:task}

We define \textbf{financial statement verification} as follows.
Given a set of financial statements $\mathcal{S} = \{S_{\text{BS}}, S_{\text{IS}}, S_{\text{CFS}}\}$ comprising a balance sheet, income statement, and cash flow statement for a given fiscal year, the task is to:

\begin{enumerate}[topsep=2pt,itemsep=1pt]
    \item \textbf{Detect}: Determine whether $\mathcal{S}$ contains any numerical inconsistencies (binary classification).
    \item \textbf{Localize}: If an inconsistency exists, identify the specific line item(s) and relationship(s) that are violated.
    \item \textbf{Explain}: Provide a natural-language explanation of why the identified values are inconsistent.
\end{enumerate}

Formally, each statement $S_k$ is a set of key-value pairs $\{(l_i, v_i)\}_{i=1}^{n_k}$ where $l_i$ is a line-item label and $v_i \in \mathbb{R}$ is its numerical value.
A set of \emph{accounting constraints} $\mathcal{C} = \{c_1, \ldots, c_m\}$ defines relationships that must hold across these values (e.g., $v_{\text{Assets}} = v_{\text{Liabilities}} + v_{\text{Equity}}$).
An error is a violation of at least one constraint: $\exists\, c_j \in \mathcal{C}$ such that $c_j(\mathcal{S})$ evaluates to false.

For detection, the model must produce a prediction $\hat{y} \in \{0, 1\}$ where $y = 1$ indicates at least one constraint violation is present.
For localization, the model must output a set of line-item identifiers $\hat{L} \subseteq \mathcal{L}$ where $\mathcal{L} = \bigcup_k \{l_i : (l_i, v_i) \in S_k\}$ is the set of all line items.
For explanation, the model must produce a free-text justification $\hat{e}$ referencing the violated constraint, the specific values, and why they are inconsistent.

This three-tiered formulation captures increasing levels of verification capability, from basic anomaly detection to the kind of reasoned assessment expected of human auditors.
We distinguish two evaluation settings.
In an \emph{implicit-constraint} setting, the model receives only the statements and must infer which relationships to check from its accounting knowledge.
In a \emph{guided-checklist} setting, the prompt lists common relationships to examine.
The main cross-model results in this draft use the guided-checklist CoT setting for comparability, while zero-shot prompting represents the implicit setting in the released framework.

\subsection{Data Source: SEC 10-K Filings via XBRL}
\label{sec:data}

We source financial data from the U.S.\ Securities and Exchange Commission's (SEC) Electronic Data Gathering, Analysis, and Retrieval (EDGAR) system.
Specifically, we use the XBRL (eXtensible Business Reporting Language) structured data available through the EDGAR Company Facts API (\texttt{data.sec.gov/api/xbrl/companyfacts/}).

XBRL provides machine-readable financial data tagged with standardized US-GAAP taxonomy concepts, enabling precise extraction of line items across the three core financial statements.
We select companies from the S\&P 500 index to ensure data quality and representativeness.
From an initial set of 50 companies, we retain 43 after excluding companies whose XBRL filings use non-standard taxonomies (primarily financial holding companies) or lack key line items needed for error injection.
The retained company-year set spans FY2014--FY2026 and is concentrated in FY2024--FY2026 (41 of 43 company-years).

For each company, we extract the following financial statement components:

\paragraph{Balance Sheet.}
Total assets, total liabilities, stockholders' equity, cash and cash equivalents, accounts receivable, inventory, property/plant/equipment (net), current assets, current liabilities, long-term debt, and retained earnings.

\paragraph{Income Statement.}
Revenue, cost of goods sold, gross profit, operating expenses, operating income, interest expense, income before tax, income tax expense, and net income.

\paragraph{Cash Flow Statement.}
Net cash from operations, net cash from investing, net cash from financing, capital expenditures, depreciation and amortization, and net change in cash.

We verify the consistency of the extracted data against the original XBRL filings by checking fundamental accounting identities (e.g., $\text{Assets} = \text{Liabilities} + \text{Equity}$) and discard any companies where the source data itself contains inconsistencies due to XBRL tagging variations or restatements.

The resulting dataset covers companies with combined total assets exceeding \$17 trillion, with an average of 37 current-year line items per company (range: 20--42).
Where XBRL concepts are not directly available, we derive missing fields using fundamental accounting relationships (e.g., total liabilities $=$ total assets $-$ total equity for companies that do not report total liabilities as a separate XBRL concept).
These derived fields are constructed to be \emph{exactly} consistent, ensuring that the ``clean'' versions of our financial statements satisfy the relationships our error injection targets.

\subsection{Error Taxonomy}
\label{sec:taxonomy}

We develop a four-category, twelve-subtype taxonomy of financial statement errors grounded in auditing practice (\Cref{tab:taxonomy}; full details with targeted line items in \Cref{app:taxonomy}).
The taxonomy defines twelve subtypes, but the current benchmark instantiates eleven: CL-D\&A is specified for completeness and excluded from the main evaluation because depreciation and amortization is not represented consistently enough across XBRL filings.
The instantiated taxonomy is designed to span the space of numerical inconsistencies that can arise in financial statements, from simple arithmetic failures to subtle cross-statement linkage violations.

\begin{table}[t]
\centering
\small
\caption{Error taxonomy for financial statement verification.
The current benchmark instantiates eleven of the twelve defined subtypes; CL-D\&A is specified but excluded from the main evaluation because XBRL depreciation reporting was insufficiently consistent.}
\label{tab:taxonomy}
\begin{tabular}{p{0.13\textwidth}p{0.12\textwidth}p{0.45\textwidth}p{0.15\textwidth}}
\toprule
\textbf{Category} & \textbf{Subtype} & \textbf{Description} & \textbf{Difficulty} \\
\midrule
\multirow{2}{*}{\makecell[l]{Arithmetic\\Error (AE)}}
  & AE-Row & Line items do not sum to stated subtotal & Easy \\
  & AE-Col & Subtotals do not sum to stated total & Easy \\
\midrule
\multirow{4}{*}{\makecell[l]{Cross-Stmt\\Linkage (CL)}}
  & CL-NI/RE & Net income $\neq$ change in retained earnings (IS$\to$BS) & Medium \\
  & CL-NI/CFS & Net income $\neq$ CFS operating section start (IS$\to$CFS) & Medium \\
  & CL-Cash & CFS ending cash $\neq$ BS cash balance (CFS$\to$BS) & Medium \\
  & CL-D\&A & Depreciation inconsistent across statements (IS$\to$CFS); specified only & Excluded \\
\midrule
\multirow{2}{*}{\makecell[l]{Year-over-\\Year (YoY)}}
  & YoY-Open & Prior year ending balance $\neq$ current opening balance & Medium \\
  & YoY-Chg & Computed period change $\neq$ stated change & Medium \\
\midrule
\multirow{4}{*}{\makecell[l]{Magnitude\\(MR)}}
  & MR-Minor & Perturbation $<1\%$ of true value & Hard \\
  & MR-Mod & Perturbation 1--5\% of true value & Medium \\
  & MR-Sig & Perturbation 5--20\% of true value & Easy \\
  & MR-Ext & Perturbation $>20\%$ of true value & Easy \\
\bottomrule
\end{tabular}
\end{table}

\paragraph{Category 1: Arithmetic Errors (AE).}
These errors violate within-statement summation constraints.
An AE-Row error modifies one line item such that the items no longer sum to their stated subtotal (e.g., current assets components not summing to total current assets).
An AE-Col error modifies a subtotal such that it disagrees with the stated total (e.g., current assets plus non-current assets not equaling total assets).

\paragraph{Category 2: Cross-Statement Linkage Errors (CL).}
These errors break numerical relationships that must hold \emph{across} financial statements.
CL-NI/RE modifies net income on the income statement so it no longer matches the change in retained earnings on the balance sheet.
CL-NI/CFS introduces a discrepancy between the income statement's net income and the starting point of the operating activities section of the cash flow statement.
CL-Cash makes the ending cash balance on the cash flow statement disagree with the cash line on the balance sheet.
CL-D\&A would introduce an inconsistency in depreciation and amortization between the income statement and the cash flow statement, but is excluded from the current release for the reporting-consistency reasons above.
The instantiated cross-statement errors require models to maintain and check relationships across documents---a capability that is largely absent from prior financial QA benchmark design.

\paragraph{Category 3: Year-over-Year Consistency Errors (YoY).}
Financial statements include comparative figures from prior periods.
YoY-Open modifies a prior-year ending balance so it disagrees with the current-year opening balance (which should be identical in the absence of restatements).
YoY-Chg introduces a discrepancy between a computed year-over-year change and its components.

\paragraph{Category 4: Magnitude/Rounding Errors (MR).}
These errors test sensitivity to the \emph{size} of numerical perturbations.
We perturb a single value by a controlled percentage, creating errors ranging from minor rounding differences ($<1\%$) to extreme deviations ($>20\%$).
This category enables the magnitude sensitivity analysis that is central to our empirical contribution.

\subsection{Error Injection Methodology}
\label{sec:injection}

\begin{algorithm}[t]
\caption{Error Injection for \benchmark{}}
\label{alg:injection}
\begin{algorithmic}[1]
\Require Clean financial statements $\mathcal{S}$, error type $t \in \mathcal{T}$, magnitude $m \in \{0.5, 1, 2, 5, 10, 20\}\%$
\Ensure Modified statements $\mathcal{S}'$ with ground-truth annotation $g$
\State $\mathcal{S}' \leftarrow \text{deepcopy}(\mathcal{S})$
\State $v_{\text{target}} \leftarrow \text{SelectTarget}(t, \mathcal{S}')$  \Comment{Select line item based on error type}
\State $v_{\text{original}} \leftarrow \mathcal{S}'[v_{\text{target}}]$
\State $s \leftarrow \text{Uniform}(\{-1,+1\})$
\State $\delta \leftarrow |v_{\text{original}}| \times (m/100) \times s$
\State $v_{\text{modified}} \leftarrow v_{\text{original}} + \delta$
\State $\mathcal{S}'[v_{\text{target}}] \leftarrow v_{\text{modified}}$  \Comment{Optionally round to statement precision}
\State $g \leftarrow (t, v_{\text{target}}, v_{\text{original}}, v_{\text{modified}}, m)$
\State \Return $\mathcal{S}', g$
\end{algorithmic}
\end{algorithm}

Given a clean set of financial statements $\mathcal{S}$ verified for internal consistency, we inject exactly one error per instance using the procedure in \Cref{alg:injection}.
The injection targets a specific line item determined by the error type, perturbs it by the specified magnitude, and records the ground truth for evaluation.
Using $|v_{\text{original}}|$ makes the perturbation magnitude independent of accounting sign convention: expenses, losses, cash outflows, and contra-equity accounts may be negative in the rendered statements, but the injected deviation is defined as a percentage of absolute reported magnitude.
Zero, missing, or non-numeric candidate fields are skipped for that subtype; if no valid target exists, the candidate instance is not generated.

Several design decisions merit discussion:

\paragraph{Single-error injection.}
We inject exactly one error per instance to enable unambiguous attribution of detection (or failure) to a specific error type and magnitude.
Multi-error variants are left for future work.

\paragraph{Magnitude selection.}
We use six magnitude levels: 0.5\%, 1\%, 2\%, 5\%, 10\%, and 20\%.
These span from rounding-level noise (0.5\%) to obviously material misstatements (20\%).
The 5\% threshold is commonly discussed as a preliminary quantitative rule of thumb in materiality assessments, but SEC Staff Accounting Bulletin No.\ 99 cautions that materiality cannot be determined by a numerical threshold alone and requires qualitative judgment~\citep{sec1999sab99}.
This makes the 0.5--5\% range interesting as a sensitivity region rather than as a bright-line materiality boundary.

\paragraph{Rounding.}
The generated corpus stores the original \emph{unrounded} perturbations, where the perturbed value can retain full precision.
This may produce fractional values (e.g., \$534.66M in a statement otherwise reported in whole millions), creating a detectable format anomaly.
For the LLM diagnostic sample, we additionally construct a \emph{rounded} variant in which injected values are rounded to the statement's reporting precision (typically millions or thousands of dollars).
We compare the rounded and unrounded sample variants to disentangle genuine arithmetic verification from format-based anomaly detection (\Cref{sec:analysis:strategies}).

\paragraph{Sign randomization.}
The direction of perturbation (increase or decrease) is randomized to prevent models from exploiting systematic biases.

\subsection{Benchmark Statistics}
\label{sec:stats}

The generated \benchmark{} corpus comprises \textbf{1,985 instances} drawn from \textbf{43 S\&P 500 companies} across 7 sectors (technology, consumer, healthcare, industrial, financial, telecom, and energy).
Seven companies from the initial 50 were excluded due to non-standard reporting structures (e.g., financial holding companies using different XBRL taxonomies).
For each company, we generate:
\begin{itemize}[topsep=2pt,itemsep=1pt]
    \item 1 clean (consistent) instance (43 total)
    \item 7 instantiated non-magnitude error subtypes $\times$ 6 magnitudes = 42 error-injected instances per company (the twelfth defined subtype, CL-D\&A, is excluded from the current benchmark due to inconsistent depreciation reporting across XBRL filings)
    \item 4 magnitude-category instances (MR-Minor through MR-Ext, at their defined magnitudes)
\end{itemize}

\noindent This yields $43 \times (1 + 42 + 4) = 2{,}021$ candidate instances.
After excluding instances where a specific error subtype could not be injected for a given company (primarily CL-Ending Cash, which covers 86\% of companies), the generated count is 1,985.
Because binary verification is underdetermined when the perturbed field is not rendered, we additionally label 119 generated positive instances as \emph{not enough information} and exclude them from binary metrics.
The resulting observable binary subset contains 1,866 instances (43 clean and 1,823 observable errors); the shared LLM diagnostic subset contains 105 observable binary instances after excluding 3 hidden-field positives.
\Cref{tab:dataset_stats} summarizes the generated corpus and observable evaluation subset.

\begin{table}[t]
\centering
\small
\caption{Dataset composition of \benchmark{}.
Percentages in the first block are relative to the generated corpus (1,985 instances).
The observable binary subset relabels 119 hidden-field positives as not-enough-information and excludes them from binary detection metrics.}
\label{tab:dataset_stats}
\begin{tabular}{lrr}
\toprule
\textbf{Metric} & \textbf{Count} & \textbf{\%} \\
\midrule
Total instances & 1,985 & 100.0 \\
Clean instances & 43 & 2.2 \\
Error instances & 1,942 & 97.8 \\
Not-enough-information positives & 119 & 6.0 \\
Observable binary subset & 1,866 & 94.0 \\
\quad Observable clean & 43 & 2.2 \\
\quad Observable errors & 1,823 & 91.8 \\
\midrule
\multicolumn{3}{l}{\emph{By generated error category}} \\
\quad Arithmetic (AE) & 516 & 26.6 \\
\quad Cross-Statement (CL) & 738 & 38.0 \\
\quad Year-over-Year (YoY) & 516 & 26.6 \\
\quad Magnitude (MR) & 172 & 8.9 \\
\midrule
\multicolumn{3}{l}{\emph{By generated magnitude range}} \\
\quad $<1\%$ & 340 & 17.5 \\
\quad $1$--$5\%$ & 631 & 32.5 \\
\quad $5$--$20\%$ & 633 & 32.6 \\
\quad $>20\%$ & 338 & 17.4 \\
\midrule
\multicolumn{3}{l}{\emph{By difficulty}} \\
\quad Easy & 301 & 15.2 \\
\quad Moderate & 1,297 & 65.3 \\
\quad Hard & 301 & 15.2 \\
\quad Very Hard & 43 & 2.2 \\
\bottomrule
\end{tabular}
\end{table}

Each instance includes the three financial statements formatted as they would appear in a filing, a verification prompt, and comprehensive ground-truth annotations.
The generated corpus contains 97.8\% error instances and 2.2\% clean instances; consequently, we report FPR, precision, and recall rather than relying solely on accuracy, since a trivial ``always flag as erroneous'' baseline would achieve 97.8\% accuracy on the generated corpus.
For binary LLM evaluation, we use the observable 105-instance diagnostic subset with a higher proportion of clean instances (41.0\%) to enable meaningful FPR estimation.
We also release a class-balanced observable diagnostic split of 86 instances (43 clean, 43 errors) for calibration analyses that should not depend on the generated corpus prevalence.
The magnitude distribution is approximately uniform across the four ranges, enabling robust analysis of detection sensitivity.

\subsection{Evaluation Metrics}
\label{sec:metrics}

We evaluate models on three levels corresponding to the task formulation in \Cref{sec:task}:

\paragraph{Detection.}
Binary classification: does the model correctly identify whether the statements contain an error?
We report accuracy, precision, recall, and F1 score.
We also report the false positive rate (FPR) on clean instances, as a high FPR would render the system impractical for deployment.

\paragraph{Localization.}
Among correctly detected errors, does the model identify the correct line item(s)?
We define localization accuracy as the fraction of detected errors for which the model's identified location overlaps with the ground-truth error location.

\paragraph{Explanation quality.}
For qualitative analysis, we define a 3-point scale for explanation evaluation:
(1)~\emph{correct}: explanation correctly identifies the violated relationship and references appropriate values;
(2)~\emph{partially correct}: explanation identifies the general area of the error but misstates values or relationships;
(3)~\emph{incorrect}: explanation is wrong or fabricated.
We report qualitative observations on explanation patterns; systematic explanation evaluation with automated metrics is deferred to future work.

\paragraph{Magnitude sensitivity curve.}
For each model and error type, we plot the detection rate as a function of error magnitude, producing a \emph{psychometric-style} sensitivity curve.
The magnitude at which detection exceeds 50\% (the \emph{detection threshold}, $m_{50}$) provides a scalar summary of model sensitivity.

\paragraph{Confidence intervals.}
Because we evaluate on finite samples, we compute 95\% Wilson score confidence intervals~\citep{wilson1927} for detection rates and report them when interpreting boundary estimates such as 0\% FPR.
The Wilson interval is preferred over the Wald interval for proportions near 0 or 1, as it provides correct coverage even at boundary values.

\section{Experimental Setup}
\label{sec:setup}

\subsection{Baselines and Models}
\label{sec:models}

We evaluate \benchmark{} with a deterministic rule-based verifier and fourteen complete frontier-LLM runs; one additional attempted run (Gemini 2.5 Pro through OpenRouter) is reported separately as incomplete because of gateway failures.

\paragraph{Rule-based verifier.}
We implement a deterministic verifier that programmatically checks all accounting relationships encoded in the error taxonomy: within-statement arithmetic sums (11 checks), cross-statement linkages (4 checks), and year-over-year consistency.
This system represents a ceiling for structured verification---it has perfect recall by construction when the violated relationship is among its checked rules, but cannot distinguish injected errors from inherent noise in real financial data.

\paragraph{Evaluated LLMs.}
The complete-run model set spans multiple access paths and both closed-source and open-weight systems.
It includes Claude Sonnet 4~\citep{claude2025}, Claude Opus 4.6, Claude Sonnet 4.6, GPT-4.1~\citep{openai2025gpt41}, GPT-5.5, GPT-5.4, GPT-5.2, DeepSeek V3.2~\citep{deepseek2025v3}, DeepSeek R1~\citep{deepseek2025r1}, Qwen 3 235B~\citep{qwen2025qwen3}, Llama 4 Maverick and Llama 4 Scout~\citep{meta2025llama4}, Gemma 3 27B~\citep{google2025gemma3}, and MiniMax M2.5~\citep{minimax2025}.
This selection supports both cross-provider comparison and within-family/access-path analysis for Anthropic, OpenAI, DeepSeek, Meta, and Google/Gemma; the incomplete Gemini run is documented in \Cref{app:repro} but excluded from main capability comparisons.

\paragraph{Sampling.}
The original provider-API LLM sample contains 108 instances: all 43 clean instances plus 65 error-injected instances, with two instances per error-type/magnitude cell wherever possible.
Before computing binary metrics, we remove the 3 positive instances whose perturbed field is not rendered to the model, yielding the main 105-instance observable diagnostic subset (43 clean, 62 error-injected).
The later Codex CLI GPT evaluations were run directly on this 105-instance observable subset.
The removed instances are labeled not-enough-information in the release metadata rather than treated as detectable errors.

\paragraph{Evaluation protocol.}
Provider-API models are evaluated at temperature 0 for deterministic output; the Codex CLI GPT runs use the CLI's default decoding because \texttt{codex exec} does not expose a temperature flag, with JSON output constrained by an explicit response schema.
The main comparison uses a shared guided-checklist chain-of-thought verification prompt for all complete model runs (\Cref{app:prompts}), instructing the model to check common arithmetic, cross-statement, and year-over-year relationships step by step.
For Codex CLI rows, this same verification prompt is preceded by a short access-path preamble instructing the agent not to inspect files, run commands, browse, or use tools, and to answer from the financial statements in the prompt only.
The reported CoT prompt also includes the same rounding-tolerance instruction for all models: small rounding differences within one unit of reporting precision should not be flagged as errors.
Because the prompt lists simplified subtotal relationships, we treat the main prompt as an \emph{original guided-checklist} condition rather than as a prompt-invariant measurement of financial-verification capability.
\Cref{app:prompts} therefore also specifies a \emph{completeness-aware} CoT condition that instructs models not to assume visible component line items exhaust a subtotal unless the statement explicitly indicates completeness.
The cross-model table therefore reports a fixed-prompt diagnostic condition, not a prompt-invariant model capability claim; the completeness-aware prompt is released to support a future prompt-factorial evaluation.
Supplementary zero-shot and few-shot runs for MiniMax M2.5 produce the same qualitative always-error behavior as its CoT run, but these pilot runs are not used for cross-model comparisons.
All strategies instruct the model to respond in structured JSON format with detection decision, error location, and explanation.
Responses are parsed with a multi-stage pipeline: JSON extraction from code fences, inline JSON matching, and keyword heuristic fallback.

\subsection{Implementation}
\label{sec:implementation}

All experiments are implemented in Python.
Financial data is fetched from the SEC EDGAR XBRL API with appropriate rate limiting (10 requests/second) and User-Agent identification as required by SEC policy.
Error injection follows the procedure in \Cref{alg:injection}.

Models are accessed via their respective provider APIs or provider CLIs, with exact access paths recorded in \Cref{app:repro}.
The evaluation is fully deterministic: each instance is evaluated exactly once, with no repeated sampling or majority voting.
Exact model IDs, access paths, parsing details, and known response failures are reported in \Cref{app:repro}.
The complete benchmark construction pipeline, evaluation framework, and analysis code are publicly available.
Evaluation of the provider-API 108-instance stratified sample takes approximately 10--15 minutes per model at a cost of approximately \$1--5 USD depending on the provider and model; the Codex CLI GPT runs on the 105-instance observable subset took roughly 30--35 minutes per model in our environment.

\section{Results and Analysis}
\label{sec:results}

We separate two result types.
The \emph{realistic rounded} condition is the more conservative estimate of substantive verification ability, because it removes visible decimal-format artifacts.
The \emph{unrounded} condition is retained as a diagnostic artifact-sensitivity setting, because the complete cross-model runs share that common input rendering.
Accordingly, we avoid leaderboard claims and use the cross-model table to study calibration, prompt sensitivity, and construct validity under a fixed original guided-checklist prompt.

\subsection{Rule-Based Verification Baseline}
\label{sec:results:baseline}

We evaluate the rule-based verifier across a range of detection thresholds $\tau$ (\Cref{tab:baseline}).
The verifier checks 15 accounting relationships that are guaranteed consistent in our clean data by construction (e.g., $\text{Assets} = \text{Liabilities} + \text{Equity}$, IS net income $=$ CFS net income).

\begin{table}[t]
\centering
\small
\caption{Rule-based verifier performance on \benchmark{} at different detection thresholds $\tau$.
The verifier achieves 0\% FPR but limited recall: even at the tightest threshold, 47\% of injected errors go undetected because they violate relationships not among the 15 explicitly checked.}
\label{tab:baseline}
\begin{tabular}{rccccr}
\toprule
$\bm{\tau}$ & \textbf{Acc.} & \textbf{Prec.} & \textbf{Recall} & \textbf{F1} & \textbf{FPR} \\
\midrule
0.01\% & 53.8 & 100.0 & 52.8 & 69.1 & 0.0 \\
0.1\%  & 53.5 & 100.0 & 52.5 & 68.8 & 0.0 \\
1.0\%  & 40.0 & 100.0 & 38.6 & 55.7 & 0.0 \\
5.0\%  & 23.2 & 100.0 & 21.5 & 35.4 & 0.0 \\
10.0\% & 14.9 & 100.0 & 13.0 & 23.0 & 0.0 \\
20.0\% & 5.8  & 100.0 & 3.7  & 7.1  & 0.0 \\
\bottomrule
\end{tabular}
\end{table}

Three key findings emerge.
First, the rule-based verifier achieves \textbf{perfect precision (0\% FPR)} at all thresholds when restricted to relationships that hold exactly in the clean data.
Second, even at the tightest threshold ($\tau = 0.01\%$), \textbf{recall is only 52.8\%}---meaning 47\% of injected errors go entirely undetected.
Third, detection varies dramatically by error category: \textbf{cross-statement linkage errors are detected at 100\%} (because the verifier explicitly checks IS$\leftrightarrow$CFS and CFS$\leftrightarrow$BS linkages), but arithmetic errors at only 31\%, magnitude errors at 32\%, and year-over-year errors at 14.2\%.
Among correctly detected errors, \textbf{localization accuracy is 62.6\%}---the verifier identifies the correct problematic field in roughly two-thirds of cases.

An important corollary: at tight thresholds, detection depends on \emph{which field} was perturbed, not \emph{by how much}.
Detection rates across magnitude bins ($<1\%$, $1$--$5\%$, $5$--$20\%$, $>20\%$) are approximately equal at $\tau=0.01\%$, because the verifier either checks a given relationship (and catches any deviation) or does not check it (and misses the error regardless of size).
Only at higher thresholds ($\tau \geq 1\%$) do small-magnitude errors begin to pass under the threshold, creating the expected magnitude dependence.

\begin{figure}[t]
\centering
\includegraphics[width=0.95\textwidth]{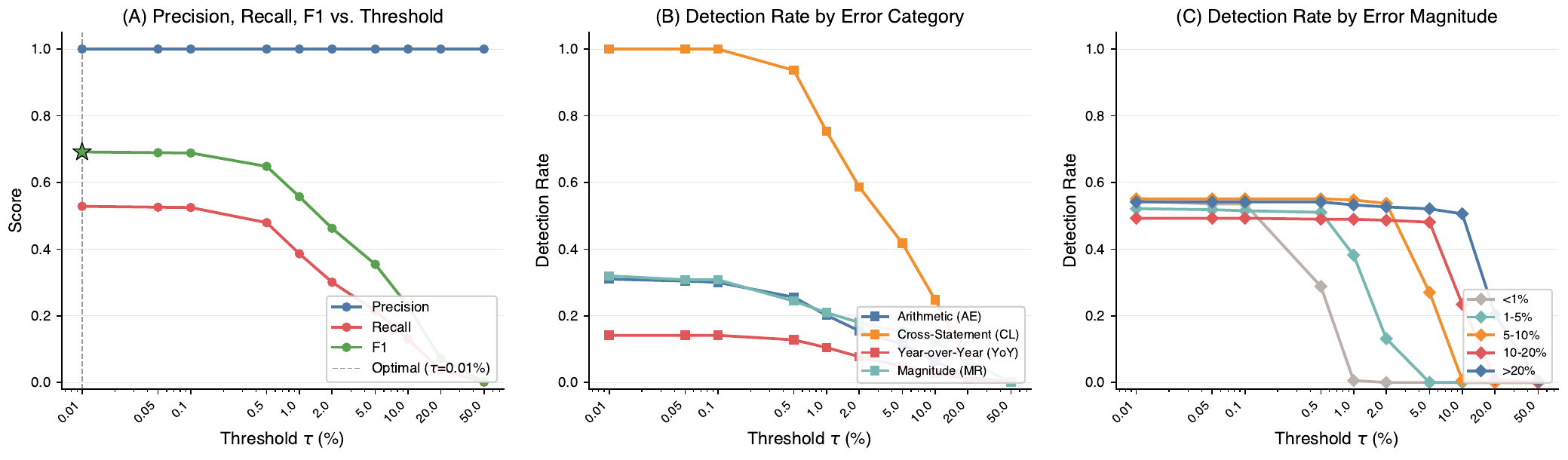}
\caption{Detection performance of the rule-based verifier as a function of threshold $\tau$.
Recall degrades monotonically with increasing threshold, while precision remains at 100\% across all thresholds.
The shaded region marks a 1--5\% sensitivity band often used as a preliminary quantitative rule-of-thumb region; materiality ultimately requires qualitative judgment under SAB 99.}
\label{fig:threshold}
\end{figure}

\begin{figure}[t]
\centering
\includegraphics[width=0.85\textwidth]{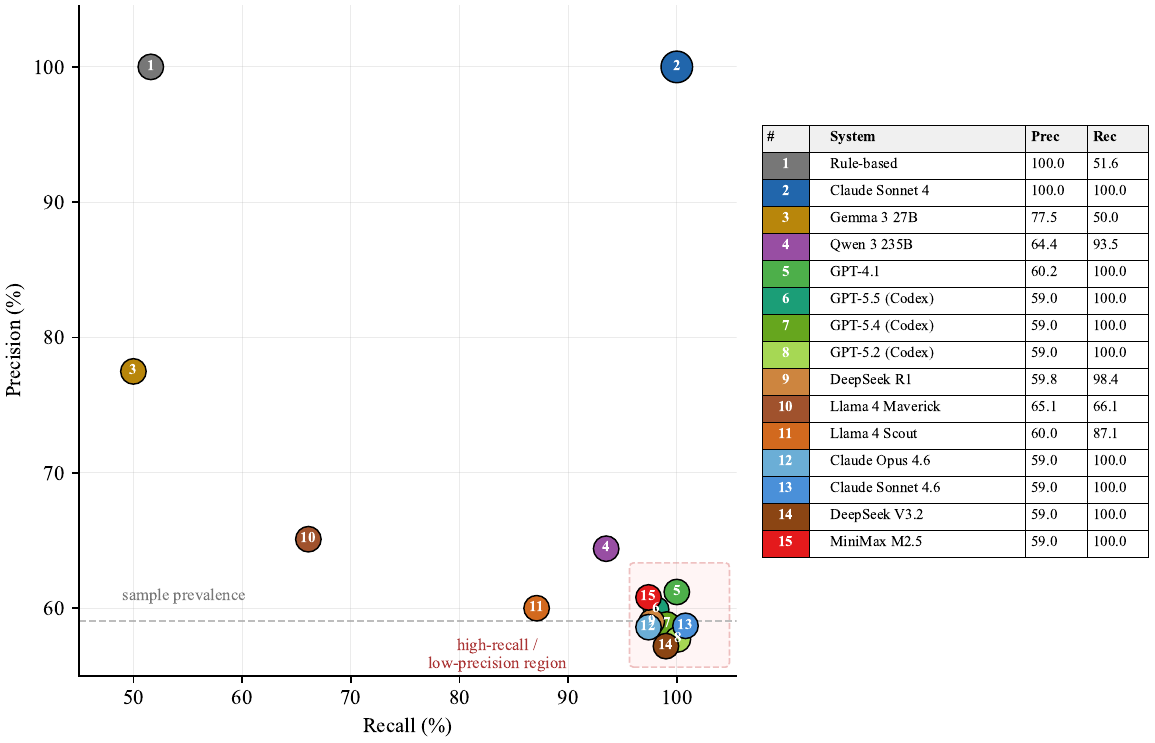}
\caption{Precision-recall trade-off across verification systems on the 105-instance observable unrounded diagnostic subset.
Numbered markers avoid label overlap; the adjacent legend table gives exact precision and recall values.
The shaded region highlights the high-recall/low-precision operating point associated with error-detection bias under the original guided-checklist prompt.}
\label{fig:precision_recall}
\end{figure}

\Cref{tab:per_category} and \Cref{fig:threshold} present these results in detail.
\Cref{fig:precision_recall} provides a precision-recall visualization that contextualizes the rule-based verifier alongside the LLM results presented in \Cref{sec:results:llm}.

\begin{table}[t]
\centering
\small
\caption{Rule-based verifier detection rate by error category at $\tau=0.01\%$.
Cross-statement errors are fully detected because the verifier explicitly checks those linkages; other categories have partial coverage.}
\label{tab:per_category}
\begin{tabular}{lrrrr}
\toprule
\textbf{Category} & \textbf{Instances} & \textbf{Detected} & \textbf{Rate (\%)} & \textbf{Loc.\ Acc.\ (\%)} \\
\midrule
Arithmetic (AE)       & 516 & 160 & 31.0 & 58.1 \\
Cross-Statement (CL)  & 738 & 738 & 100.0 & 68.3 \\
Year-over-Year (YoY)  & 516 &  73 & 14.2 & 41.1 \\
Magnitude (MR)        & 172 &  55 & 32.0 & 63.6 \\
\midrule
\textbf{Overall}      & 1,942 & 1,026 & 52.8 & 62.6 \\
\bottomrule
\end{tabular}
\end{table}

\subsection{The Verification Gap}
\label{sec:results:noise}

The 47\% of errors that the rule-based verifier misses constitute the \emph{verification gap}---the space of accounting constraints that are not explicitly programmed but that a knowledgeable verifier should check.
This gap arises from two sources:

\paragraph{Partial observability.}
Real financial statements contain dozens of line items per section, with many not captured by standard XBRL concept mappings.
For instance, ``Cash from Investing Activities'' includes capital expenditures, acquisitions, divestitures, and investment sales---but our XBRL extraction captures only capital expenditures.
A rule that checks ``CapEx $=$ Cash from Investing'' would produce discrepancies exceeding 1,000\% on clean data.
We found that when we extend the verifier to check 30 relationships (including these partial-observability rules), recall increases to 100\% but FPR also increases to 100\%---a precision-recall trade-off that cannot be resolved without contextual understanding.

\paragraph{Implicit constraints.}
Some consistency checks require domain knowledge not expressed in the statement structure itself.
For example, verifying that retained earnings changed by exactly (net income $-$ dividends $-$ share repurchases) requires knowing the accounting relationship between these items---knowledge that is implicit rather than explicit in the financial statements.

This analysis has important implications.
The verification gap quantifies the value that a context-aware system (such as an LLM with accounting knowledge) could add over a purely rule-based approach: the ability to check the approximately 47\% of constraints that require either reasoning about incomplete data or implicit domain knowledge.
\benchmark{} is designed to test exactly this capability.

\subsection{Dataset Analysis}
\label{sec:results:dataset}

The benchmark provides systematic coverage across all four error categories (\Cref{tab:dataset_stats}), with up to 46 error-injected instances per company spanning 7 non-magnitude error subtypes at 6 magnitude levels (42) plus 4 magnitude-category instances.
\begin{figure}[t]
\centering
\includegraphics[width=0.9\textwidth]{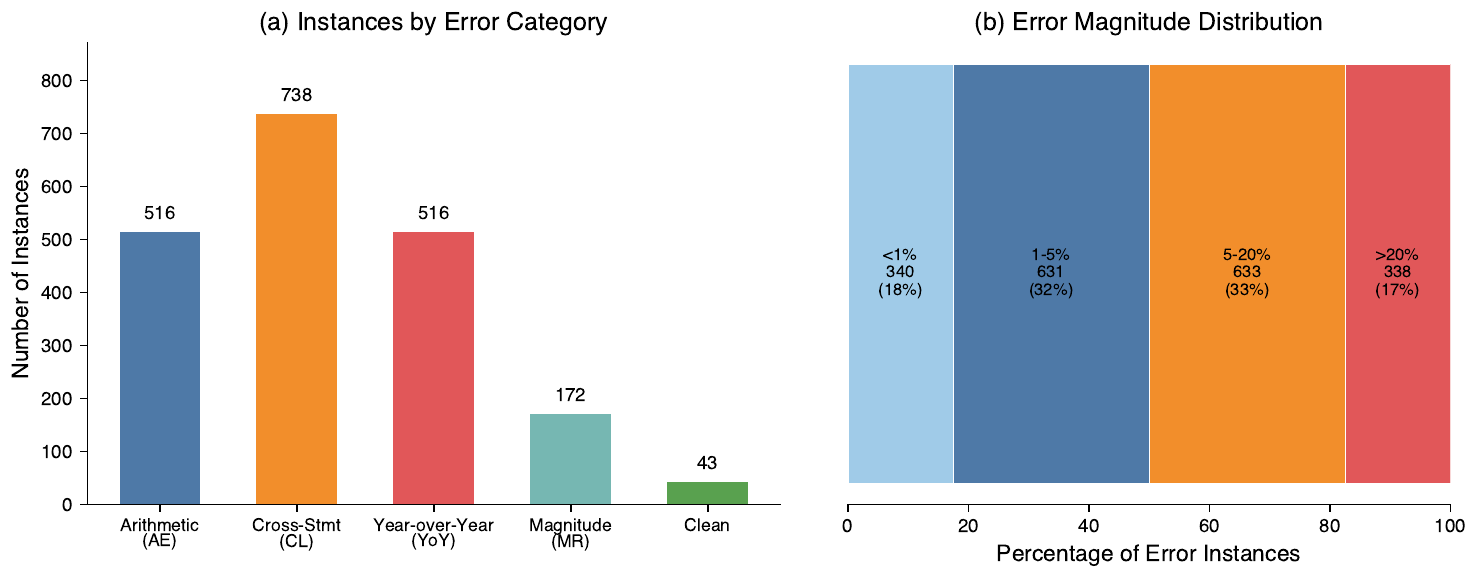}
\caption{Dataset composition of \benchmark{}.
Left: number of instances by error category (AE = arithmetic, CL = cross-statement, YoY = year-over-year, MR = magnitude/rounding).
Right: distribution of error magnitudes across the benchmark.}
\label{fig:dataset_composition}
\end{figure}

\begin{figure}[t]
\centering
\includegraphics[width=0.85\textwidth]{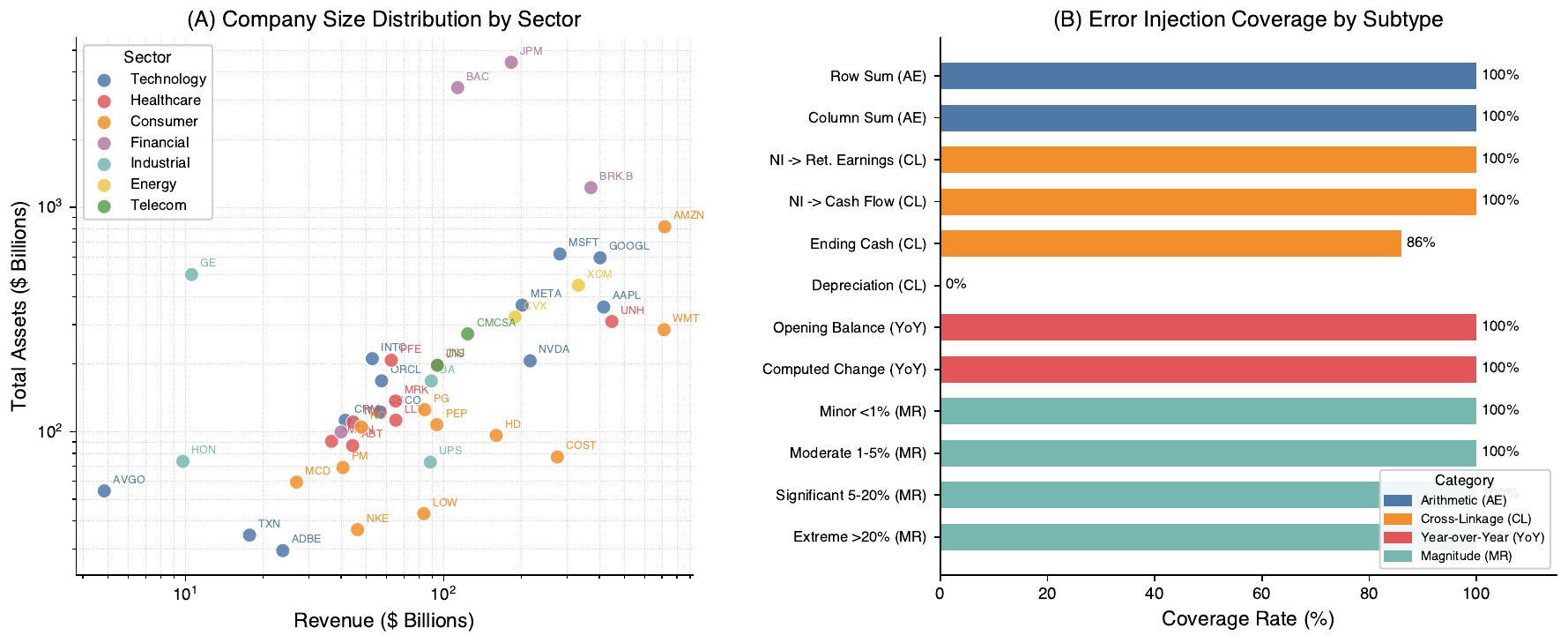}
\caption{Left: Revenue vs.\ total assets for the 43 companies in \benchmark{} (log-log scale, colored by sector), showing the wide range of company sizes and financial profiles covered.
Right: Error injection coverage rate by error subtype; ten of the eleven instantiated subtypes achieve 100\% coverage.}
\label{fig:diversity}
\end{figure}

\Cref{fig:dataset_composition} shows the distribution of instances across error categories and magnitudes.
The 43 companies span 7 sectors: technology (12), consumer (11), healthcare (8), industrial (4), financial (4), telecom (2), and energy (2) (\Cref{fig:diversity}).
This sectoral diversity ensures that the benchmark captures variation in financial reporting practices across industries.

The companies span a wide range of financial profiles: revenue from \$4.8B to \$716.9B (median \$84.3B) and total assets from \$29.5B to \$4.4T (median \$136.9B), with an average of 12.2 checkable accounting relationships per company (\Cref{fig:diversity}).
All 43 companies have complete data for all three financial statements and prior-year comparatives, ensuring full coverage of year-over-year error types.
Ten of the eleven instantiated error subtypes achieve 100\% injection coverage; CL-Ending Cash covers 86\% (37/43).
The twelfth defined subtype, CL-D\&A, is excluded from the main benchmark because depreciation and amortization is not reported consistently enough across the relevant XBRL concepts.

\begin{figure}[t]
\centering
\includegraphics[width=0.95\textwidth]{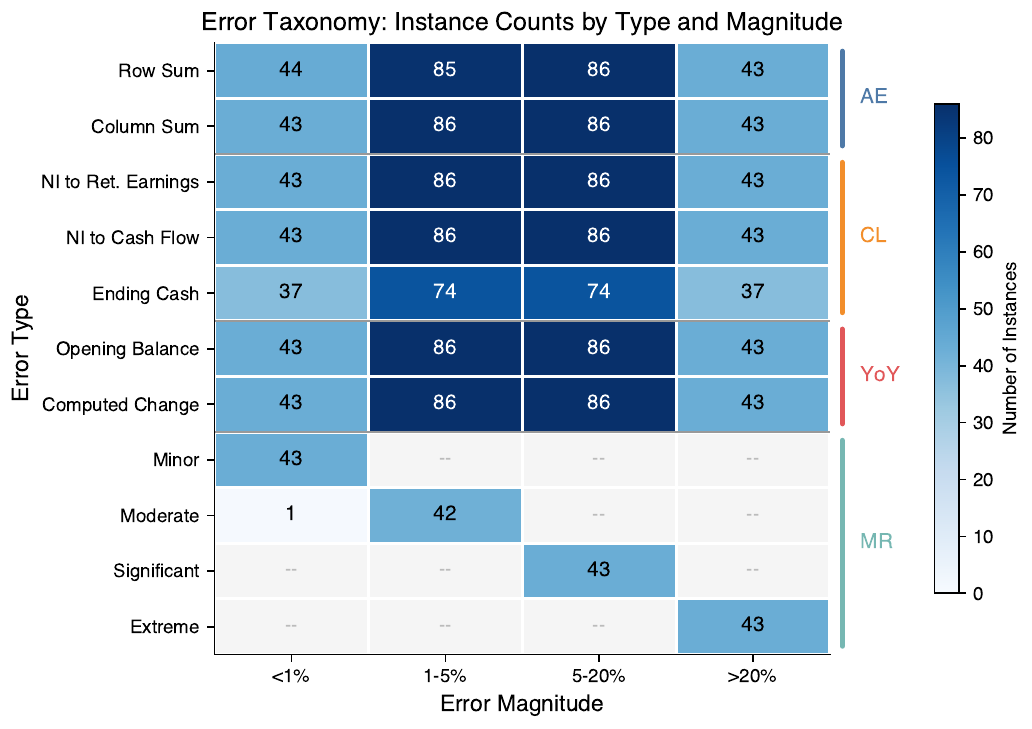}
\caption{Heatmap of benchmark instances by error type (rows) and magnitude level (columns). Each cell shows the number of instances, enabling systematic analysis of detection sensitivity across the full error type $\times$ magnitude grid.}
\label{fig:taxonomy_heatmap}
\end{figure}

The error type $\times$ magnitude grid (\Cref{fig:taxonomy_heatmap}) confirms uniform coverage across cells, enabling fine-grained sensitivity analysis.

\subsection{Benchmark Validation}
\label{sec:results:validation}

We validate the benchmark through several checks:

\paragraph{Error injection fidelity.}
For each injected error, we verify that (a) exactly one line item was modified, (b) the modification matches the specified magnitude within rounding tolerance, and (c) the ground truth correctly records the original and modified values.
All 1,942 generated error instances pass these checks.

\paragraph{Clean instance consistency.}
Clean instances are verified against the fundamental accounting identity ($\text{Assets} = \text{Liabilities} + \text{Equity}$) and inter-statement linkages that we explicitly construct (IS net income $=$ CFS net income; CFS ending cash $=$ BS cash).
All 43 clean instances satisfy these core constraints.

\paragraph{Error discriminability and observability.}
We verify that error-injected instances are distinguishable from clean instances at their injected error location.
The minimum absolute perturbation across the unrounded instances is \$0.5M (at 0.5\% magnitude for small line items).
Because realistic rounded rendering can remove visible decimal cues, we treat rounded detection as the more conservative estimate of substantive verification ability.
We additionally check whether the perturbed field is actually rendered in the formatted statement.
The 119 generated positives that fail this check are labeled \emph{not enough information} and excluded from all binary detection metrics; in the 108-instance LLM sample, this removes 3 positives and leaves 62 observable errors.

\subsection{Framework for LLM Evaluation}
\label{sec:results:framework}

\benchmark{} includes a complete evaluation framework for LLM-based verification.
We provide prompt templates for the original evaluated condition and for follow-up ablations (\Cref{app:prompts}):

\paragraph{Zero-shot.}
The model receives financial statements and a general verification instruction.
This tests whether models can identify inconsistencies without explicit guidance on what to check.

\paragraph{Chain-of-thought (CoT).}
The model is instructed to systematically check specific accounting relationships step by step, mirroring audit procedures.
This tests whether structured reasoning improves detection, particularly for cross-statement errors that require maintaining relationships across document sections.
The framework also includes zero-shot, few-shot, and completeness-aware variants used for limited supplementary runs or released for follow-up prompt-factorial evaluation.

\subsection{LLM Evaluation on the Observable Unrounded Diagnostic Sample}
\label{sec:results:llm}

\Cref{tab:llm_results} presents results from the original guided-checklist prompt on the 105-instance observable unrounded diagnostic subset.
This subset is the shared 108-instance LLM sample after relabeling three hidden-field positives as not-enough-information and excluding them from binary metrics.
The rule-based row is recomputed on the same 105 observable instances; its full-corpus performance is reported separately in \Cref{tab:baseline}.
The attempted Gemini 2.5 Pro run is omitted from the main table because 40/108 calls failed through the gateway (\Cref{app:repro}).

\begin{table}[t]
\centering
\small
\caption{Observable-subset comparison under the original guided-checklist prompt on the shared unrounded diagnostic sample.
The table is a fixed-prompt diagnostic comparison, not a prompt-invariant leaderboard.
$n=105$ for all rows: 43 clean statements and 62 observable error statements after excluding 3 not-enough-information positives from the attempted 108-instance sample.
Codex CLI rows use the same verification checklist plus a no-tools access-path preamble and JSON output schema.}
\label{tab:llm_results}
\begin{tabular}{lcccccc}
\toprule
\textbf{System} & $\bm{n}$ & \textbf{Acc.} & \textbf{Prec.} & \textbf{Rec.} & \textbf{F1} & \textbf{FPR} \\
\midrule
Rule-based ($\tau=0.01\%$) & 105 & 71.4 & \textbf{100.0} & 51.6 & 68.1 & \textbf{0.0} \\
\midrule
Claude Sonnet 4 & 105 & \textbf{100.0} & \textbf{100.0} & \textbf{100.0} & \textbf{100.0} & \textbf{0.0} \\
Gemma 3 27B & 105 & 61.9 & 77.5 & 50.0 & 60.8 & 20.9 \\
Qwen 3 235B & 105 & 65.7 & 64.4 & 93.5 & 76.3 & 74.4 \\
GPT-4.1 & 105 & 61.0 & 60.2 & \textbf{100.0} & 75.2 & 95.3 \\
GPT-5.5 (Codex CLI) & 105 & 59.0 & 59.0 & \textbf{100.0} & 74.3 & 100.0 \\
GPT-5.4 (Codex CLI) & 105 & 59.0 & 59.0 & \textbf{100.0} & 74.3 & 100.0 \\
GPT-5.2 (Codex CLI) & 105 & 59.0 & 59.0 & \textbf{100.0} & 74.3 & 100.0 \\
DeepSeek R1 & 105 & 60.0 & 59.8 & 98.4 & 74.4 & 95.3 \\
Claude Opus 4.6 & 105 & 59.0 & 59.0 & \textbf{100.0} & 74.3 & 100.0 \\
Claude Sonnet 4.6 & 105 & 59.0 & 59.0 & \textbf{100.0} & 74.3 & 100.0 \\
DeepSeek V3.2 & 105 & 59.0 & 59.0 & \textbf{100.0} & 74.3 & 100.0 \\
Llama 4 Maverick & 105 & 59.0 & 65.1 & 66.1 & 65.6 & 51.2 \\
Llama 4 Scout & 105 & 58.1 & 60.0 & 87.1 & 71.1 & 83.7 \\
MiniMax M2.5 & 105 & 59.0 & 59.0 & \textbf{100.0} & 74.3 & 100.0 \\
\bottomrule
\end{tabular}
\end{table}

\begin{figure}[t]
\centering
\includegraphics[width=0.95\textwidth]{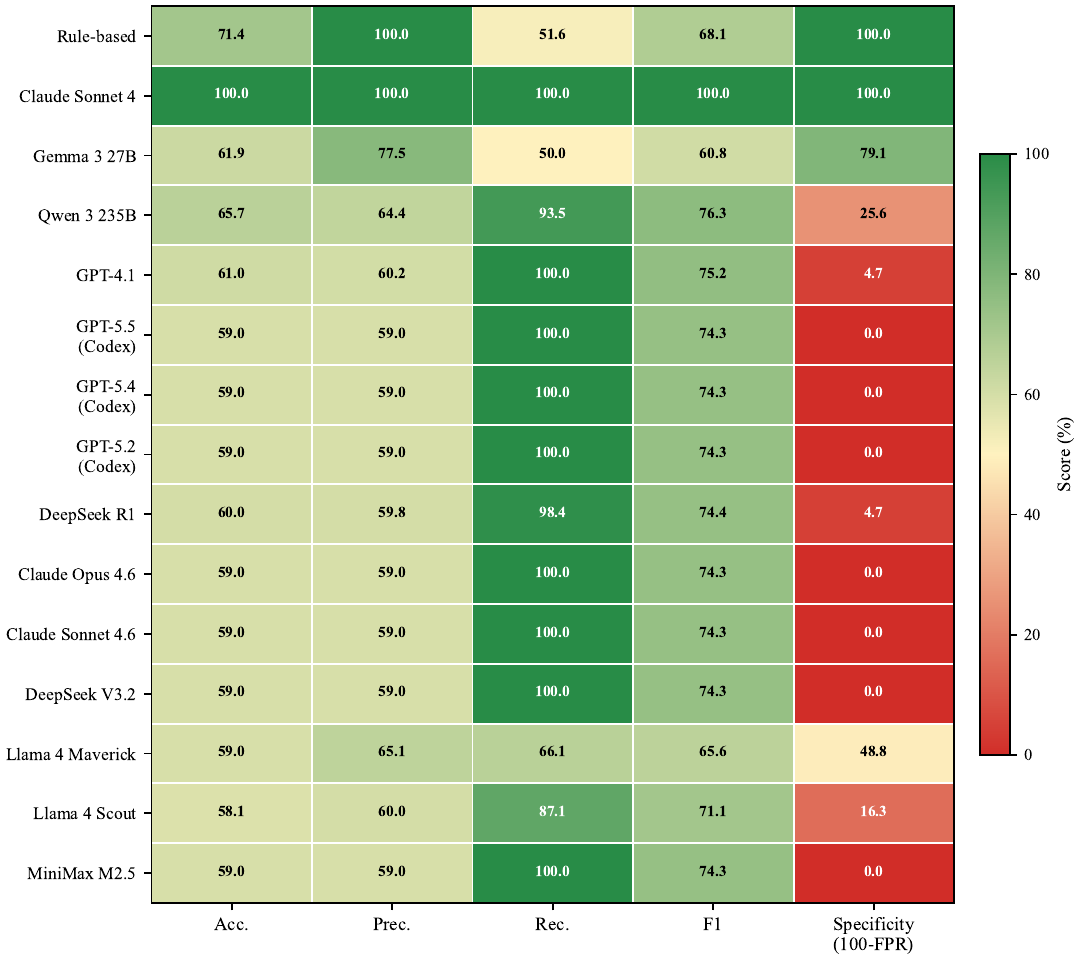}
\caption{Heatmap comparison of complete verification systems across five metrics on the 105-instance observable unrounded diagnostic subset.
High recall is common, but specificity on clean statements varies from 0\% to 100\%.}
\label{fig:model_comparison}
\end{figure}

\paragraph{Finding 1: under the original guided checklist, the central observed challenge is calibration rather than detection.}
Claude Sonnet 4 correctly identifies all 62 observable injected errors and produces no false alarms on the 43 clean instances.
At the opposite extreme, seven complete runs---GPT-5.5, GPT-5.4, GPT-5.2, Claude Opus 4.6, Claude Sonnet 4.6, DeepSeek V3.2, and MiniMax M2.5---flag every clean instance as erroneous.
GPT-4.1 and DeepSeek R1 are only slightly less extreme, each misclassifying 41 of 43 clean examples (95.3\% FPR).
Between these poles, Gemma 3 27B, Llama 4 Maverick, Qwen 3 235B, and Llama 4 Scout occupy progressively less selective operating points, with false positive rates of 20.9\%, 51.2\%, 74.4\%, and 83.7\%, respectively.
Thus nine of fourteen complete LLM runs exhibit severe error-detection bias under this prompt, while the 0--100\% FPR spread shows that clean-instance calibration, not raw arithmetic sensitivity alone, drives measured usefulness.
\Cref{fig:precision_recall} and \Cref{fig:model_comparison} visualize this calibration spectrum.

\paragraph{Finding 2: error-category recall and clean specificity must be read together.}
\Cref{tab:category_results} shows that many systems achieve high recall on observable injected errors only by sacrificing clean specificity.
The seven always-error runs achieve 100\% recall in every error category but 0\% clean specificity.
Claude Sonnet 4 and the rule-based verifier both achieve 100\% clean specificity; they differ in recall, with Claude detecting all observable error categories in this unrounded diagnostic setting and the rule-based verifier missing errors outside its programmed checks.

\begin{table}[t]
\centering
\footnotesize
\caption{Detection rate (\%) by error category on the observable unrounded diagnostic subset, with clean specificity in the bottom row.
``Biased'' aggregates the seven always-error runs: GPT-5.5, GPT-5.4, GPT-5.2, Claude Opus 4.6, Claude Sonnet 4.6, DeepSeek V3.2, and MiniMax M2.5.}
\label{tab:category_results}
\begin{tabular}{lrrrrrrrrrr}
\toprule
\textbf{Cat.} & $\bm{n}$ & \textbf{Cl.S4} & \textbf{Gem3} & \textbf{Qwen} & \textbf{GPT} & \textbf{R1} & \textbf{Mav} & \textbf{Sct} & \textbf{Biased} & \textbf{Rule} \\
\midrule
AE       & 16 & 100.0 & 25.0 & 75.0 & 100.0 & 100.0 & 43.8 & 87.5 & 100.0 & 12.5 \\
CL       & 24 & 100.0 & 75.0 & 100.0 & 100.0 & 95.8 & 87.5 & 83.3 & 100.0 & 100.0 \\
YoY      & 15 & 100.0 & 33.3 & 100.0 & 100.0 & 100.0 & 60.0 & 86.7 & 100.0 & 26.7 \\
MR       &  7 & 100.0 & 57.1 & 100.0 & 100.0 & 100.0 & 57.1 & 100.0 & 100.0 & 28.6 \\
\midrule
\textbf{Overall error} & 62 & \textbf{100.0} & 50.0 & 93.5 & 100.0 & 98.4 & 66.1 & 87.1 & 100.0 & 51.6 \\
\midrule
\textbf{Clean specificity} & 43 & \textbf{100.0} & 79.1 & 25.6 & 4.7 & 4.7 & 48.8 & 16.3 & 0.0 & \textbf{100.0} \\
\bottomrule
\end{tabular}
\end{table}

\paragraph{Finding 3: unrounded observable errors are detected across magnitudes, but this is a diagnostic artifact-sensitivity result.}
\Cref{tab:magnitude_results} presents detection rates by magnitude after excluding hidden-field positives.
Claude Sonnet 4 detects every observable unrounded error across all magnitude buckets.
This should not be interpreted as realistic sub-materiality verification because the unrounded rendering can expose decimal-format artifacts; \Cref{sec:analysis:strategies} quantifies the drop under rounded rendering.

\begin{table}[t]
\centering
\small
\caption{Detection rate (\%) by error magnitude on observable unrounded errors.
Several high-recall models are omitted because their curves are flat by construction: they flag every instance as erroneous.}
\label{tab:magnitude_results}
\begin{tabular}{lrrrr}
\toprule
\textbf{Magnitude} & $\bm{n}$ & \textbf{Cl. Sonnet 4} & \textbf{Qwen 3} & \textbf{Rule} \\
\midrule
$<1\%$       & 16 & 100.0 & 100.0 & 62.5 \\
$1$--$5\%$  & 16 & 100.0 &  87.5 & 43.8 \\
$5$--$20\%$ & 15 & 100.0 &  93.3 & 53.3 \\
$>20\%$     & 15 & 100.0 &  93.3 & 46.7 \\
\bottomrule
\end{tabular}
\end{table}

\begin{figure}[t]
\centering
\includegraphics[width=0.86\textwidth]{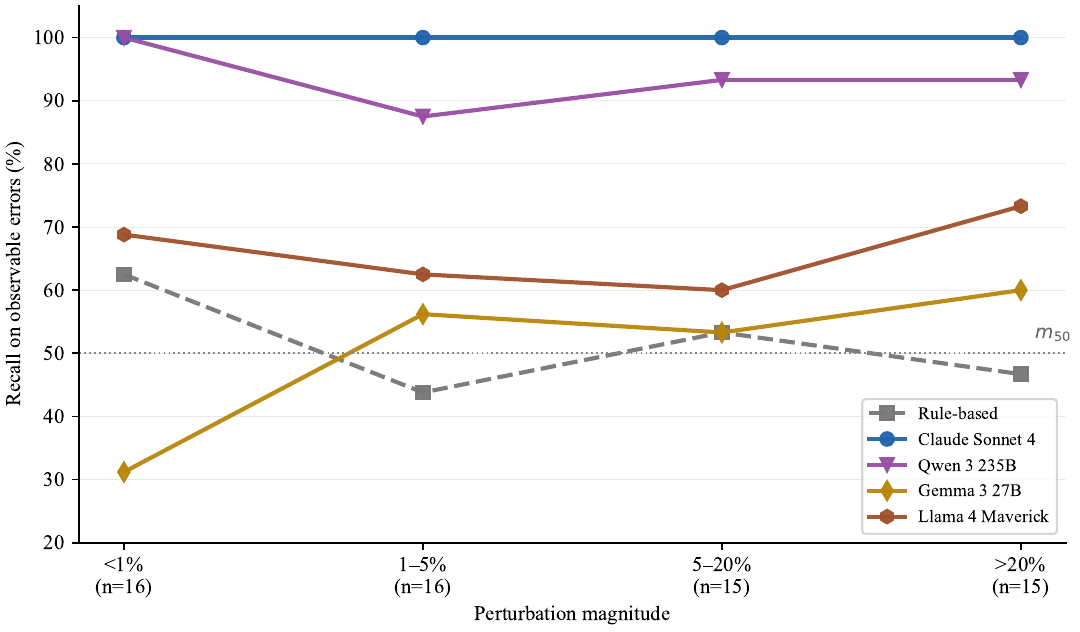}
\caption{Detection sensitivity by error magnitude for selected systems on observable unrounded errors.
Always-error systems are excluded because they do not provide meaningful sensitivity curves.}
\label{fig:magnitude_curves}
\end{figure}

\paragraph{Finding 4: the false-positive mechanism is epistemic, not arithmetic.}
Nine of fourteen complete LLM runs exhibit severe error-detection bias under the original guided-checklist prompt.
This pattern should be interpreted together with the prompt design: the checklist includes simplified subtotal relationships and can encourage models to treat omitted line items as evidence of inconsistency.
We therefore do not claim that the observed FPRs are prompt-invariant.
MiniMax shows the same always-error behavior under supplementary zero-shot and few-shot prompting, but those runs cover only one model and do not replace a cross-model prompt ablation.
Manual inspection indicates that false positives are usually arithmetically correct but semantically miscalibrated: models compute real discrepancies over visible line items in simplified statements and over-interpret those discrepancies as accounting errors.

\begin{figure}[t]
\centering
\includegraphics[width=0.98\textwidth]{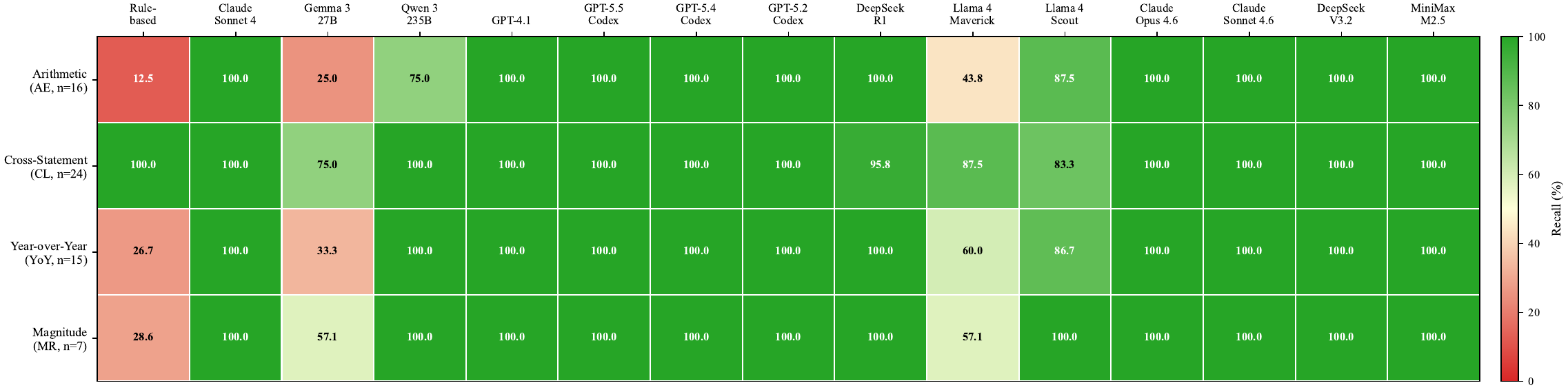}
\caption{Error-category recall on the observable unrounded diagnostic subset.
The heatmap separates error recall from clean-statement calibration, avoiding the misleading impression that high recall alone implies useful verification.}
\label{fig:category_recall}
\end{figure}

\begin{figure}[t]
\centering
\includegraphics[width=0.98\textwidth]{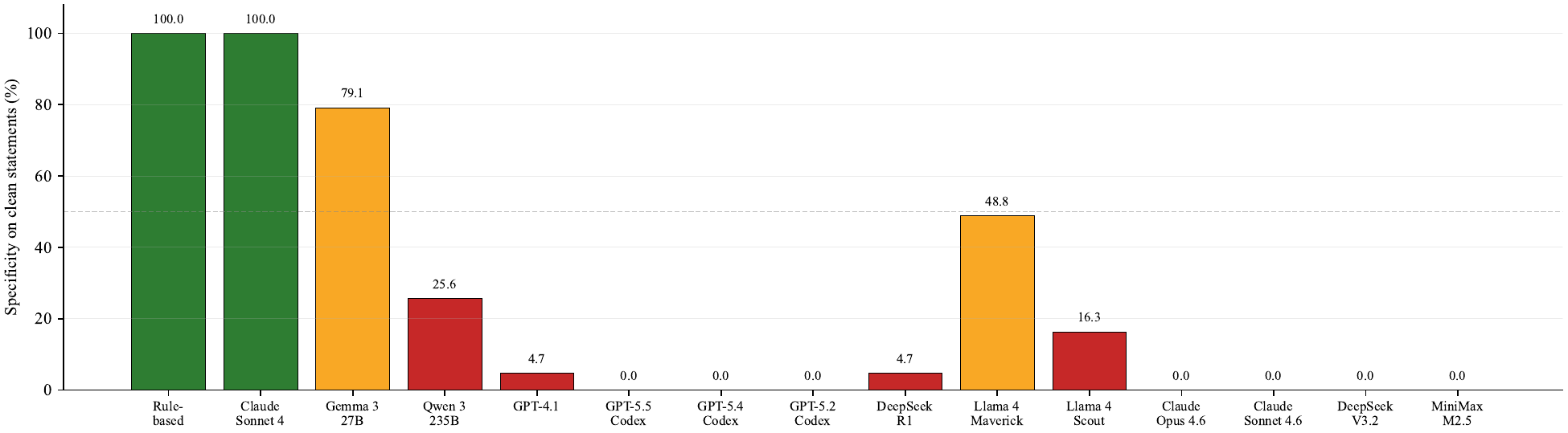}
\caption{Specificity on the 43 clean statements.
This clean split is small (Wilson 95\% upper bound 8.2\% for an observed 0\% FPR), but it exposes a large calibration spread under the original guided-checklist prompt.}
\label{fig:clean_specificity}
\end{figure}

\paragraph{Implications and caveats.}
These results reveal that financial verification performance varies widely across frontier LLMs even when evaluated under identical inputs and prompting.
The paired comparison between Claude Sonnet 4 and GPT-4.1 is particularly informative within the original guided-checklist condition: both are evaluated on the same 105 observable instances under the same prompt.
The 39-point accuracy gap (100.0\% vs. 61.0\%) and the FPR difference (0\% [95\% Wilson upper bound: 8.2\%] vs. 95.3\%) are highly significant (McNemar's $p < 10^{-9}$; \Cref{sec:analysis:significance}), but the result should be read as a paired comparison under this prompt and rendering, not as a prompt-invariant model ranking.

\section{Analysis}
\label{sec:analysis}

\subsection{What Makes Financial Verification Hard?}
\label{sec:analysis:hard}

Our benchmark construction and baseline analysis highlight several factors that make financial statement verification difficult:

\paragraph{Incomplete information.}
Real financial statements contain dozens of line items per section, but standardized XBRL taxonomies capture only a subset.
Our analysis of 43 S\&P 500 companies shows that simplified summation relationships (e.g., ``capital expenditures $=$ cash from investing'') can diverge by over 1,000\% from reality because of unmodeled components (acquisitions, divestitures, investment sales).
Any verifier---rule-based or LLM---must reason about what it \emph{cannot} see.

\paragraph{Numerical precision and rounding.}
Financial statements report figures rounded to the nearest million or thousand.
At the 0.5\% magnitude level, the absolute difference between the correct and erroneous values can be as small as \$0.5M for a \$100M line item---comparable to the rounding unit itself.
This creates an inherent ambiguity zone where verification is genuinely underdetermined.

\paragraph{Cross-document reasoning.}
Cross-statement errors (CL category) require maintaining and comparing numerical values across three separate document sections (income statement, balance sheet, cash flow statement)---a task that existing benchmarks on numerical reasoning over long documents~\citep{zhao2024docmatheval} have shown to degrade LLM performance significantly.

\paragraph{Implicit domain knowledge.}
Some verification checks require knowledge of accounting relationships that are not stated in the input.
For example, detecting a CL-NI/RE error requires knowing that net income should flow to retained earnings minus dividends---a fundamental accounting relationship but one that is implicit rather than explicitly stated in the financial statements.
The chain-of-thought prompt partially addresses this by making key relationships explicit, but the set of possible relationships is large and domain-specific.

\subsection{The Precision-Recall Spectrum in Financial Verification}
\label{sec:analysis:rulevllm}

Under the original guided-checklist prompt on the observable unrounded diagnostic subset, the complete LLM runs fall into five broad operating regimes (plus the rule-based baseline):

\begin{itemize}[topsep=2pt,itemsep=1pt]
    \item \textbf{Rule-based (deterministic)}: By restricting checks to relationships that hold exactly in clean data, the verifier achieves 0\% FPR but only 51.6\% recall on the observable diagnostic subset---missing errors that violate unchecked relationships.
    \item \textbf{Severely biased (9 models)}: GPT-5.5, GPT-5.4, GPT-5.2, Claude Opus 4.6, Claude Sonnet 4.6, DeepSeek V3.2, MiniMax M2.5, DeepSeek R1, and GPT-4.1 all achieve 95--100\% FPR, identifying nearly every clean financial statement as inconsistent.
    GPT-4.1 and DeepSeek R1 each correctly identify 2 of 43 clean instances; the other seven flag every clean instance as erroneous.
    \item \textbf{Near-biased and partially calibrated}: Llama 4 Scout achieves 87.1\% recall with 83.7\% FPR, while Qwen 3 235B achieves 93.5\% recall with 74.4\% FPR.
    Both maintain high recall but still generate many false alarms.
    \item \textbf{Moderate/conservative}: Llama 4 Maverick (66.1\% recall, 51.2\% FPR) and Gemma 3 27B (50.0\% recall, 20.9\% FPR) trade recall for specificity, suggesting narrower or more cautious verification behavior.
    \item \textbf{Claude Sonnet 4 (calibrated in this diagnostic condition)}: Achieves 100\% precision and 100\% recall on the observable unrounded subset, successfully distinguishing injected observable errors from expected clean-statement discrepancies under this prompt.
\end{itemize}

The contrast across the fourteen complete LLM runs is the central diagnostic result.
All complete models process the same financial statements under the same original guided-checklist verification prompt family, yet only Claude Sonnet 4 correctly classifies all 43 clean instances.
With nine of fourteen complete runs exhibiting 95--100\% FPR---including multiple runs from the same provider families as calibrated or less-biased runs---severe error-detection bias appears common in this fixed-prompt setting.
A plausible interpretation is that flagging everything as suspicious is a ``safe'' response when a model is asked to look for errors, whereas calibrated behavior requires tolerating expected imprecision and omitted line items.

\subsection{Statistical Significance and Paired Analysis}
\label{sec:analysis:significance}

Since Claude Sonnet 4 and GPT-4.1 were evaluated on the \emph{same} observable unrounded instances under the \emph{same} original guided-checklist prompt, we apply McNemar's test for paired nominal data~\citep{mcnemar1947} to assess whether the performance difference is statistically significant in that condition.

\Cref{tab:mcnemar} presents the $2 \times 2$ contingency table of correct/incorrect predictions.
Of 105 observable binary instances, 64 are concordant correct (both models correct on 62 observable error instances and on 2 clean instances), 41 are discordant in Claude's favor (Claude correct, GPT-4.1 wrong---all clean instances that GPT-4.1 false-alarms on), and none are discordant in GPT-4.1's favor after the hidden positives are removed from binary scoring.

\begin{table}[t]
\centering
\small
\caption{McNemar's test: Claude Sonnet 4 vs. GPT-4.1 on the same 105 observable unrounded instances under the original guided-checklist prompt.
The 41:0 discordant split is highly significant ($p < 10^{-9}$), driven by GPT-4.1's false positives on clean instances.}
\label{tab:mcnemar}
\begin{tabular}{lcc}
\toprule
 & \textbf{GPT-4.1 Correct} & \textbf{GPT-4.1 Wrong} \\
\midrule
\textbf{Claude Correct} & 64 & 41 \\
\textbf{Claude Wrong} & 0 & 0 \\
\bottomrule
\end{tabular}
\vspace{4pt}

\raggedright\scriptsize
McNemar $\chi^2$ (continuity-corrected) $= 39.02$, $p = 4.19 \times 10^{-10}$.
Exact binomial $p = 9.09 \times 10^{-13}$.
\end{table}

This quantitative decomposition supports a narrower claim than a global leaderboard: on the 105-instance unrounded diagnostic sample under the original guided-checklist prompt, Claude Sonnet 4 significantly outperforms GPT-4.1, and the gap is entirely attributable to clean-instance specificity.
On the 43 clean instances, all 41 discordant pairs favor Claude; on the 62 observable error instances, both models detect every error.
Thus, under this prompt and rendering, the central observed challenge is calibration rather than detection.

\paragraph{Pairwise analysis across complete runs.}
Extending McNemar's test to the $\binom{14}{2} = 91$ complete-run pairs on the same observable unrounded sample (Bonferroni-corrected $\alpha = 0.05/91 = 0.00055$) shows the same qualitative pattern: Claude Sonnet 4 significantly outperforms every other complete LLM run in this fixed diagnostic condition, while most non-Claude pairwise differences do not survive correction.
Seven models (GPT-5.5, GPT-5.4, GPT-5.2, Claude Opus 4.6, Claude Sonnet 4.6, DeepSeek V3.2, and MiniMax M2.5) produce identical always-error predictions on the observable subset ($\kappa = 1.0$).
We report these tests as paired diagnostics for one prompt/rendering condition, not as a prompt-invariant ranking of providers or model families.

\subsection{Anatomy of False Positives: Why Models Miscalibrate}
\label{sec:analysis:fp}

To understand the root cause of GPT-4.1's 95.3\% FPR, we categorize the explanations from its 41 false positive responses on clean instances.
We find that \textbf{95.1\% (39/41)} of false positives flag \emph{real but benign} numerical discrepancies---the model's arithmetic is correct, but it misinterprets missing line items as errors.
The remaining 2 instances (4.9\%) involve structural differences in financial reporting.

\Cref{tab:fp_patterns} shows the distribution of false positive patterns.

\begin{table}[t]
\centering
\small
\caption{Categorization of GPT-4.1's 41 false positive explanations on clean instances.
The dominant failure mode is computing subtotals from visible line items in simplified statements, which omit items that the reported totals include.}
\label{tab:fp_patterns}
\begin{tabular}{lrr}
\toprule
\textbf{FP Pattern} & \textbf{Count} & \textbf{\%} \\
\midrule
Operating income formula mismatch & 11 & 26.8 \\
Net income derivation mismatch & 9 & 22.0 \\
Retained earnings rollforward & 6 & 14.6 \\
Balance sheet equation & 5 & 12.2 \\
Incomplete subtotal summation & 5 & 12.2 \\
Revenue/COGS structure & 3 & 7.3 \\
Other & 2 & 4.9 \\
\bottomrule
\end{tabular}
\end{table}

The fundamental failure mode is that GPT-4.1 \textbf{treats simplified financial statements as if they were complete}.
For example, on Cisco's clean instance, GPT-4.1 correctly computes $\text{Cash} + \text{AR} + \text{Inventory} = \$18.2\text{B}$ but then flags this as inconsistent with Total Current Assets of \$35.0B---the \$16.8B difference being unlisted current assets (prepaid expenses, deferred tax assets, etc.) that are omitted in the simplified format.
Similarly, on Amazon's clean instance, GPT-4.1 computes $\text{IBT} - \text{Tax} = \$60.9\text{B} \neq \text{Net Income of } \$77.7\text{B}$, not recognizing that equity method income and other items bridge the gap.

The 2 true negatives (UPS and Berkshire Hathaway) succeed not because GPT-4.1 uses a different verification strategy, but because those companies' visible line items happen to approximately reconcile under simple arithmetic.
Notably, Berkshire Hathaway's response explicitly acknowledges that the statements are ``internally consistent \emph{within the level of detail given}''---a qualification that is absent from all 41 false positive responses.

This analysis reveals that the calibration challenge is fundamentally about \emph{epistemic awareness}: a calibrated verifier must recognize the boundaries of the information presented and avoid treating absence of evidence as evidence of error.
It also reveals a prompt-design threat: an incomplete checklist can induce false positives if the model interprets each listed relationship as complete rather than conditional on the visible line items.
Claude demonstrates more of this capability under the original prompt; GPT-4.1 generally does not.

\subsection{Realistic Rounded Benchmark and Diagnostic Unrounded Variant}
\label{sec:analysis:strategies}

The cross-model results above are useful as a diagnostic stress test, but they should not be treated as the primary estimate of realistic verification performance because the original injected values can contain visible decimal artifacts.
Manual inspection of Claude Sonnet 4's true-positive explanations reveals two recurring behaviors.
In many cross-statement cases, the model performs genuine arithmetic verification (e.g., checking whether net income matches across the income statement and cash flow statement, or whether ending cash on the cash flow statement equals the balance-sheet cash line).
In other cases, the explanation explicitly points to fractional or ``suspicious'' decimal values that stand out in statements otherwise reported in whole millions or thousands.
Because these cues often co-occur in the same response, we avoid treating free-text keyword counts as a primary quantitative result and instead measure the artifact directly through a rendering ablation.

\paragraph{Arithmetic verification.}
For cross-statement linkage errors (CL), Claude frequently performs genuine arithmetic checks, computing Net Income on the IS and CFS independently and reporting the discrepancy (e.g., ``Net Income on CFS (103,868.64) does not match Net Income on IS (101,832)'').
These explanations show that at least part of Claude's performance reflects real constraint checking rather than superficial anomaly spotting.

\paragraph{Anomaly detection via decimal patterns.}
A significant fraction of detections involve identifying values with ``unusual decimal'' patterns in statements otherwise reported in whole numbers.
For example, when a value is perturbed by 0.5\% from \$532M to \$534.66M, Claude flags the ``.66'' decimal as anomalous.
This artifact arises because error injection operates on raw values, producing fractional amounts that would not appear in legitimate rounded financial statements.

\paragraph{Rounded benchmark estimate.}
The cleanest way to quantify this confound is to remove the decimal cue.
We therefore round injected errors in the observable diagnostic sample to the statement's reporting precision where rounding changes the rendered value (whole millions or thousands, depending on the statement) and re-evaluate Claude Sonnet 4 on the same observable binary subset.
\Cref{tab:rounded_ablation} shows the results.

\begin{table}[t]
\centering
\footnotesize
\caption{Realistic rounded benchmark versus unrounded diagnostic variant for Claude Sonnet 4 on the 105-instance observable subset.
Rounding injected errors to statement precision removes visible decimal-format cues.
Recall drops 21.0 percentage points, while precision remains perfect.}
\label{tab:rounded_ablation}
\begin{tabular}{lccccc}
\toprule
\textbf{Condition} & \textbf{Acc.} & \textbf{Prec.} & \textbf{Rec.} & \textbf{F1} & \textbf{FPR} \\
\midrule
Rounded realistic rendering & 87.6 & 100.0 & 79.0 & 88.3 & 0.0 \\
Unrounded diagnostic rendering & 100.0 & 100.0 & 100.0 & 100.0 & 0.0 \\
\midrule
$\Delta$ (rounded $-$ unrounded) & $-$12.4 & 0.0 & $-$21.0 & $-$11.7 & 0.0 \\
\bottomrule
\end{tabular}
\end{table}

Three findings emerge from this analysis.
First, recall drops from 100.0\% to 79.0\% ($-$21.0pp), confirming that decimal-format artifacts substantially inflate detection rates in the unrounded diagnostic rendering.
Second, \textbf{precision remains at 100\%} and \textbf{FPR at 0\%}: the main effect of rounding is on recall rather than specificity for this model.
Third, partial rounded checks for Qwen 3 235B and Llama 4 Maverick reinforce that rendering effects are model-dependent: Qwen drops from 93.5\% to 83.9\% recall with unchanged 74.4\% FPR on the observable subset, while Llama 4 Maverick remains at 66.1\% recall and 51.2\% FPR.
Because rounded results are not available for every complete run, we restrict rounded-rendering claims to the models evaluated under that rendering and use the unrounded cross-model table only as an artifact-sensitivity diagnostic.

\subsection{Relationship to Existing Benchmarks}
\label{sec:analysis:comparison}

An important question is whether performance on existing financial QA benchmarks predicts verification performance.
We do not measure this directly in the present study, so we treat orthogonality as a hypothesis rather than as an established result.
Two structural differences motivate that hypothesis:

\begin{enumerate}[topsep=2pt,itemsep=1pt]
    \item \textbf{Task asymmetry}: QA requires generating correct answers; verification requires assessing whether given relationships hold---distinct operations~\citep{mathcomp2025}.
    \item \textbf{Calibration as a unique dimension}: Verification requires distinguishing genuine errors from expected discrepancies---a capability not tested by QA benchmarks, where all provided data is assumed correct.
\end{enumerate}

The large spread in verification behavior among models that are all marketed as frontier or near-frontier systems suggests that verification \emph{may} constitute an orthogonal evaluation dimension, but direct correlation with FinanceBench, FinQA, TAT-QA, or a comparable financial QA/math benchmark is needed before making a stronger claim.

\section{Discussion}
\label{sec:discussion}

\subsection{Implications for Financial AI Deployment}

Our benchmark design and baseline analysis have direct implications for the deployment of LLMs in financial applications:

\paragraph{Verification as a safety check.}
LLMs are increasingly used to generate financial analyses and summaries~\citep{lin2026openfinllm}.
Our results suggest that LLM-assisted verification may be feasible in some settings, but the capability varies dramatically---even among leading models and even before considering prompt and rendering effects.
Claude Sonnet 4's 79.0\% recall with 0\% observed FPR on the rounded observable sample is promising but far from exhaustive verification, while the 95--100\% FPR exhibited by nine of fourteen complete runs on the unrounded diagnostic sample shows that na\"ive deployment of many frontier models would produce an overwhelming number of false alarms.
That nine of fourteen complete runs---including multiple runs from providers represented elsewhere in the table---exhibit this failure mode underscores the importance of task-specific, version-specific evaluation.

\paragraph{Materiality thresholds.}
The 5\% threshold is best understood as a preliminary quantitative rule of thumb, not a safe harbor: SAB 99 explicitly cautions against exclusive reliance on any numerical threshold and emphasizes qualitative factors~\citep{sec1999sab99}.
Our unrounded diagnostic results show that Claude Sonnet 4 detects all observable perturbations below 5\% (\Cref{tab:magnitude_results}), but the rounded analysis shows that this sensitivity can be inflated by rendering artifacts.
The practical implication is therefore cautious: an LLM verifier may be useful as a pre-screening tool for potential misstatements, but any materiality assessment still requires human professional judgment.

\paragraph{The calibration challenge.}
Under the original guided-checklist prompt, the contrast across complete LLM runs suggests that the key observed challenge is not error detection per se (twelve of fourteen complete runs achieve $\geq$87\% recall on observable unrounded errors) but \emph{calibration}---the ability to distinguish genuine errors from expected discrepancies in real financial data.
Claude Sonnet 4's 0\% observed false positive rate demonstrates that this calibration is achievable on the current sample, while the 95--100\% FPR exhibited by nine of fourteen complete runs shows that it is not yet typical under this prompting and rendering condition.
This finding aligns with recent work on neuro-symbolic approaches~\citep{venra2026} that argues for constraining LLM arithmetic responsibility, though our false-positive analysis suggests that arithmetic tools alone will not solve the epistemic problem of omitted line items.
The question of \emph{why} Claude is the only complete run in our sample to achieve this level of calibration---whether through training data composition, alignment, prompting interaction, or access-path defaults---remains an important open question.

\paragraph{Intra-family calibration gap.}
Our evaluation of three Anthropic-family runs reveals a notable \emph{intra-family calibration gap}: Claude Sonnet 4 achieves 0\% FPR, while Claude Opus 4.6 and Claude Sonnet 4.6 both achieve 100\% FPR, each on the same observable instances under the original guided-checklist prompt.
This shows that calibration cannot be assumed to transfer across model versions or access defaults, even within the same provider family.
The false positive patterns of Opus 4.6 and Sonnet 4.6 are substantively similar to those of GPT-4.1 (treating simplified statements as complete)---the same failure mode, not a novel one.
Task-specific re-evaluation is required whenever the model, access path, or thinking configuration changes.

\paragraph{Tool-augmented verification.}
An important question is whether tool augmentation (e.g., calculator access, code generation) would mitigate the error detection bias we observe.
Our false positive analysis (\Cref{sec:analysis:fp}) suggests it would not: the biased models perform \emph{correct arithmetic} on the visible line items---the problem is not computational error but rather the failure to recognize that simplified statements omit line items that account for the discrepancy.
A calculator would confirm the arithmetic discrepancy but would not resolve the epistemic question of whether the discrepancy reflects an error or an expected consequence of abbreviation.
This distinguishes financial verification from mathematical problem-solving: the challenge is domain-aware judgment under incomplete information, not raw computation.

\paragraph{The incomplete-model problem.}
Our noise floor analysis reveals that automated verification must contend with the fact that standard data sources (XBRL) provide only a partial view of financial statement structure.
Any deployed system must be robust to this partial observability---a requirement that differentiates financial verification from numerical reasoning in controlled settings~\citep{tang2025financereasoning}.
Claude's success in the unrounded diagnostic condition suggests that some frontier LLMs may have internalized useful accounting-domain priors for incomplete financial data, but the rounded ablation shows that part of the measured sensitivity can come from surface-level decimal anomalies (\Cref{sec:analysis:strategies}).

\subsection{Limitations and Threats to Validity}
\label{sec:limitations}

\paragraph{Internal validity.}
The complete LLM comparison uses an observable diagnostic subset of 105 binary instances (43 clean, 62 error-injected) rather than the full generated corpus.
While stratification ensures coverage of all error-type/magnitude cells, the small per-cell counts limit our ability to assess within-cell variance.
For boundary estimates, uncertainty remains meaningful: an observed 0\% FPR over 43 clean instances has a 95\% Wilson upper bound of 8.2\%.
We evaluate at temperature 0 for deterministic output, so stochastic variation across runs is not a factor, but results may differ at non-zero temperatures.

\paragraph{Construct validity.}
Our benchmark injects single, isolated errors via random perturbation.
Real financial misstatements often involve multiple coordinated errors (e.g., inflating revenue while proportionally adjusting costs to maintain plausible margins).
Our single-error design simplifies the task relative to real-world fraud detection and may overestimate practical verification capability.
Additionally, unrounded error injection produces fractional values (e.g., \$534.66M) in statements otherwise reported in whole millions, creating a ``decimal anomaly'' signal.
The rounded analysis in \Cref{sec:analysis:strategies} shows that removing this cue reduces Claude Sonnet 4's observable-subset recall from 100.0\% to 79.0\%, meaning that the unrounded diagnostic setting overestimates arithmetic verification capability on properly rounded data.
We therefore frame the unrounded cross-model table as an artifact-sensitivity diagnostic and restrict rounded-rendering claims to the subset of models evaluated under rounded rendering.
The original CoT prompt lists simplified subtotal checks and may itself induce false positives when a model treats an incomplete checklist as complete; the completeness-aware prompt in \Cref{app:prompts} is designed to test this threat, but we do not make prompt-invariant claims.
We also present financial statements as formatted text rather than the complex layouts (PDFs with tables, footnotes, charts) used in actual filings.

\paragraph{Observability.}
A generated positive instance whose perturbed field is not visible to the model is underdetermined for the binary task.
We now label the 119 generated hidden-field positives (including 3 in the attempted LLM sample) as not-enough-information and exclude them from binary metrics.
This fixes the evaluated binary benchmark but also documents an important design constraint for future releases: error injection should either render the perturbed field or explicitly use a three-way label space.

\paragraph{External validity.}
We report fourteen complete frontier-LLM runs plus one incomplete Gemini 2.5 Pro attempt, spanning closed-source and open-weight systems and multiple model families.
This provides broader coverage than prior verification-specific work, but results may not generalize to future releases or to different provider access paths.
All complete runs are evaluated under the same original CoT prompting family on the same observable unrounded instances, enabling direct comparison within that condition.
In supplementary experiments, we also evaluated MiniMax under zero-shot and few-shot prompting; all three strategies produced the same 100\% FPR, suggesting that MiniMax's error detection bias is not unique to the CoT wording, but this single-model result does not substitute for a cross-model prompt ablation.
We note that an earlier attempt to evaluate DeepSeek R1 via a third-party API gateway encountered an 89\% error rate; successful evaluation was achieved by switching providers, yielding the results reported above.
Our benchmark covers 43 S\&P 500 companies; smaller companies with less standardized reporting may present different challenges.
The XBRL data source inherits tagging quality limitations that may not reflect all filing formats.

\paragraph{Statistical validity.}
The clean split remains small: with 43 clean instances, a 0\% observed FPR still has a Wilson upper bound of 8.2\%.
Similarly, category-level detection rates are based on $n=7$ to $n=24$ observable errors per category, yielding wide confidence intervals.
To reduce prevalence artifacts, the release now includes a class-balanced observable diagnostic split of 86 instances (43 clean, 43 errors), but stronger calibration claims still require additional clean company-years across varied reporting granularities and line-item visibility conditions.

\paragraph{Contamination and memorization risk.}
All evaluated models are trained on large internet corpora that likely include SEC filings.
This raises the concern that verification performance may reflect memorization of specific company financials rather than genuine numerical reasoning~\citep{memorization2025}.
Our benchmark design partially mitigates this: error injection modifies the original values, so memorized ``correct'' figures could help detect perturbations but would also produce false positives on clean (unmodified) instances---a pattern observed in the severely biased models but not in the calibrated run.
Nevertheless, we cannot definitively distinguish memorization from reasoning without controlled experiments using synthetic companies or post-training-cutoff filings, which we leave to future work.

\paragraph{Robustness across these limitations.}
Despite the caveats above, the construct-validity finding remains supported: under the original guided-checklist prompt, severe error-detection bias appears in nine of fourteen complete runs, and the false-positive analysis (\Cref{sec:analysis:fp}) gives a mechanistic explanation---treating simplified statements as complete.
The rounded-versus-unrounded analysis separately shows that rendering choices can materially change measured recall.
Together, these results support the paper's main claim that financial verification benchmarks must control observability, rendering, and prompt completeness before making leaderboard-style claims.

\subsection{Future Work}
\label{sec:future}

Several extensions would strengthen this line of research:

\begin{itemize}[topsep=2pt,itemsep=1pt]
    \item \textbf{Multi-error injection}: Injecting multiple errors per instance to test detection under more realistic conditions.
    \item \textbf{Forensic error patterns}: Designing error injections that mimic real-world financial fraud patterns rather than random perturbations.
    \item \textbf{Multi-modal verification}: Extending to verification across text, tables, and charts as they appear in actual filings.
    \item \textbf{Temporal extension}: Testing verification across multiple reporting periods (quarterly sequences, annual comparisons).
    \item \textbf{Larger clean calibration split}: Adding more clean instances across multiple fiscal years, reporting granularities, and partial/complete line-item visibility beyond the released 86-instance balanced diagnostic split.
    \item \textbf{Rounded cross-model evaluation}: Running the realistic rounded rendering for every evaluated model before making any leaderboard-style ranking claims.
    \item \textbf{Explanation evaluation}: Developing automated metrics for explanation quality to enable larger-scale evaluation.
    \item \textbf{Fine-tuning for verification}: Investigating whether fine-tuning on verification-specific data can improve detection capabilities, particularly for cross-statement errors.
    \item \textbf{Broader model evaluation}: Evaluating reasoning-specialized models (o3, QwQ) and domain-specific financial LLMs to investigate what architectural or training factors enable calibrated verification.
    \item \textbf{Prompting strategy ablations}: Comparing zero-shot, original guided-checklist, completeness-aware guided-checklist, and few-shot prompting across representative models to disentangle prompt effects from model capability.
    \item \textbf{Cross-version stability}: Systematic evaluation of how verification calibration varies across model checkpoints within a provider, to understand whether calibrated behavior is stable or ephemeral.
\end{itemize}

\subsection{Ethical Considerations}
\label{sec:ethics}

\paragraph{Dual-use risk.}
Our error taxonomy and injection methodology could, in principle, inform adversarial manipulation of financial statements by revealing which error types and magnitudes are hardest for automated systems to detect.
We believe this risk is mitigated by two factors: (1)~the error types we inject (arithmetic violations, cross-statement linkage breaks) are well-known to auditors and are routinely checked in standard audit procedures; and (2)~real financial fraud typically involves \emph{coordinated} multi-statement manipulation designed to maintain apparent consistency---the opposite of our single-error injection approach.
Publishing this work enables the research community to develop more robust verification tools, which we believe outweighs the marginal dual-use risk.

\paragraph{Data and privacy.}
All financial data used in \benchmark{} is sourced from publicly available SEC EDGAR filings.
No private or non-public financial information is used.
Company names are included because they are necessary for reproducibility and the data is already public.

\paragraph{Environmental impact.}
Our evaluation requires approximately 10--15 minutes of API inference per model, representing modest computational cost relative to model training.

\section{Conclusion}
\label{sec:conclusion}

We introduced \benchmark{}, a benchmark validity study for financial statement verification---a task that sits at the intersection of financial NLP, auditing, and trustworthy AI, but is not captured by existing question-answering benchmarks.
Built from 43 S\&P 500 companies' SEC 10-K filings, \benchmark{} contains a 1,985-instance generated corpus spanning a four-category taxonomy that defines twelve subtypes and instantiates eleven in the current release, with controlled perturbation magnitudes and observability labels.

Our evaluation yields three main conclusions.
First, under the original guided-checklist prompt, financial verification is calibration-sensitive: many models detect observable injected errors, but many cannot reliably refrain from flagging clean statements.
Second, performance forms a broad calibration spectrum, ranging from 0\% observed false positives to near-universal false alarming, with substantial variation both across providers and within model families.
Third, benchmark design matters: rounded rendering shows that visible format anomalies can materially inflate measured recall in the unrounded diagnostic variant, and observability labels show that hidden-field perturbations must be removed, rendered, or treated as not-enough-information.

Taken together, these results suggest that financial statement verification is not just arithmetic; it is calibrated judgment under incomplete observability, prompt-induced assumptions, and realistic numerical rendering.
Benchmarks such as \benchmark{} are therefore necessary before deploying LLMs in verification workflows, auditing support, or financial analysis pipelines.
\ifanonymized
Code and data will be released upon acceptance.
\else
\benchmark{} and all associated code are publicly available at the repository URL specified in the camera-ready version.

\section*{Acknowledgments}
This work utilized Claude (Anthropic) for assistance with code development, data analysis, and manuscript preparation.
The author takes full responsibility for all content, including the accuracy of results, claims, and interpretations presented in this paper.

\paragraph{Disclosure.}
The author has no affiliation with Anthropic or any other AI model provider evaluated in this study.
All models were evaluated using their publicly available APIs under standard commercial terms.
The finding that Claude Sonnet 4 achieves the best calibration in our benchmark is an empirical result, not a product endorsement; the rounded-versus-unrounded analysis (\Cref{sec:analysis:strategies}) demonstrates that this result is partially attributable to a construct validity artifact in the benchmark itself.
\fi

\bibliographystyle{plainnat}
\bibliography{references}

\appendix
\section{Prompt Templates}
\label{app:prompts}

\subsection{Zero-Shot Prompt}
\begin{verbatim}
You are an expert financial auditor. Your task is to
verify whether the following financial statements are
internally consistent.

Determine whether there are any numerical errors or
inconsistencies. Only flag genuine errors where numbers
clearly do not add up. Small rounding differences within
one unit of reporting precision should not be flagged.

FINANCIAL STATEMENTS:
{formatted_statements}

Respond in JSON: {"has_error": true/false,
  "error_location": "...", "explanation": "..."}
\end{verbatim}

\subsection{Original Chain-of-Thought Prompt}
\begin{verbatim}
You are an expert financial auditor performing a
systematic verification of financial statements. Work
through each check step by step, showing your
calculations, before reaching a final conclusion.

IMPORTANT: Only flag genuine errors where numbers clearly
do not add up. Small rounding differences (within 1 unit
of the reporting precision) should NOT be flagged as
errors. Financial statements often report rounded figures,
and minor discrepancies from rounding are expected and
normal.

FINANCIAL STATEMENTS:
{formatted_statements}

STEP-BY-STEP VERIFICATION:

Step 1 -- Income Statement Arithmetic
  - Does Revenue + COGS = Gross Profit?
  - Does Gross Profit + OpEx + D&A = Operating Income?
  - Does Operating Income + Interest = Income Before Tax?
  - Does IBT + Tax Expense = Net Income?

Step 2 -- Balance Sheet Arithmetic
  - Does Cash + AR + Inventory = Total Current Assets?
  - Does Total Current Assets + PP&E = Total Assets?
  - Does Total Liabilities + Total Equity = Total Assets?

Step 3 -- Cash Flow Statement Arithmetic
  - Does NI + D&A + WC Changes = Cash from Operations?
  - Do subtotals (Ops + Inv + Fin) = Net Change in Cash?
  - Does Beginning Cash + Net Change = Ending Cash?

Step 4 -- Cross-Statement Linkages
  - Does Net Income on IS = Net Income on CFS?
  - Does Ending Cash on CFS = Cash on BS?
  - Is change in Retained Earnings = NI - Dividends?

Step 5 -- Year-over-Year Consistency
  - Do prior-year values match comparative presentation?
  - Is Beginning Cash on CFS = prior-year Cash on BS?

After completing all checks, provide your final answer.

Respond in JSON: {"has_error": true/false,
  "error_location": "...", "explanation": "..."}
\end{verbatim}

\subsection{Completeness-Aware Chain-of-Thought Prompt}
\begin{verbatim}
Use the original chain-of-thought prompt above, plus this
instruction before the checklist:

Do not assume visible component line items exhaust a
subtotal unless the statement explicitly says so. Financial
statements may omit components such as prepaid expenses,
other current assets, goodwill, equity-method income, or
other investing/financing items. Only flag a subtotal
mismatch when the relationship is explicitly complete or
when all required components are shown. If a discrepancy
could be explained by omitted components, state that the
visible information is insufficient and do not mark it as
an error.
\end{verbatim}

\subsection{Few-Shot Prompt}
The few-shot prompt uses the zero-shot instruction plus one
clean example and one cross-statement-linkage error example.
The demonstrations are released with the evaluation code; the
main diagnostic table does not use the few-shot condition.

\section{Reproducibility Details}
\label{app:repro}

\Cref{tab:model_ids} lists the model identifiers recorded in the result files.
All API-based runs used temperature $0$ and a maximum output budget of 4,096 tokens; no provider-supported seed was available.
The original CoT prompt in \Cref{app:prompts} was used for the main cross-model diagnostic table.
For models that emitted explicit \texttt{<think>...</think>} spans, the parser removed those spans before JSON extraction.
Responses were parsed by fenced-JSON extraction, inline-JSON matching, and then a keyword heuristic fallback.
Heuristic parses are reported separately from API failures: MiniMax M2.5 required 73/108 heuristic parses, DeepSeek R1 64/108, and Llama 4 Scout 1/108.
API failures are \emph{not} counted as model non-detections in the main table.
The Gemini 2.5 Pro attempt had 40/108 gateway failures through OpenRouter, so it is excluded from main capability comparisons and retained only as an incomplete run.

Two access-path caveats are important.
First, router-based identifiers are not always official provider snapshot names.
For GPT-4.1 we recorded OpenRouter's \path{openai/gpt-4.1}; OpenAI documentation also lists the dated snapshot \path{gpt-4.1-2025-04-14} as a supported model ID for fine-tuning, but this run did not independently verify whether OpenRouter served the base alias, the dated snapshot, or another compatibility mapping.\footnote{OpenAI model/snapshot documentation: \url{https://developers.openai.com/api/docs/guides/supervised-fine-tuning}.}
Second, Google's Gemini changelog states that \path{gemini-2.5-pro-preview-03-25} redirected to \path{gemini-2.5-pro} on June 26, 2025 and was scheduled for shutdown on December 2, 2025; the March 2026 OpenRouter run therefore cannot be interpreted as an official, stable evaluation of the original dated preview checkpoint.\footnote{Google Gemini API changelog: \url{https://ai.google.dev/gemini-api/docs/changelog}.}

For Claude Opus 4.6 and Claude Sonnet 4.6, the runs used Claude CLI version 2.1.76 with no explicit \texttt{thinking} parameter, manual thinking budget, or adaptive-thinking override recorded in the experiment logs.
Anthropic documents adaptive thinking as the recommended mode for Opus 4.6 and Sonnet 4.6 and notes that manual budget-based thinking is deprecated on those models; therefore these CLI runs should be interpreted as default Claude CLI behavior rather than a controlled manual-thinking condition.\footnote{Anthropic adaptive-thinking documentation: \url{https://platform.claude.com/docs/en/build-with-claude/adaptive-thinking}.}

For GPT-5.5, GPT-5.4, and GPT-5.2, we used Codex CLI 0.128.0 via \texttt{codex exec} on May 4, 2026, directly on the 105-instance observable binary subset.
These runs used \texttt{--ignore-user-config}, \texttt{--ignore-rules}, read-only sandboxing, \texttt{approval\_policy=\"never\"}, ephemeral sessions, a no-tools access-path prompt preamble, a JSON output schema, and medium reasoning effort.
The Codex CLI access path did not expose a temperature parameter, so these rows should be interpreted as Codex-CLI default decoding rather than temperature-0 API evaluations.

The observed API cost for the attempted 108-instance provider-API sample was approximately \$1--5 per model; the Codex CLI runs are logged separately with observed token counts in the released result JSON files.

\begin{table}[t]
\centering
\scriptsize
\caption{Recorded model identifiers and access paths for the original guided-checklist evaluation. Main binary metrics are computed on the 105-instance observable subset and exclude incomplete Gemini results. Codex CLI GPT rows were run directly on the 105-instance observable subset.}
\label{tab:model_ids}
\begin{tabularx}{\textwidth}{p{0.16\textwidth}p{0.32\textwidth}p{0.24\textwidth}X}
\toprule
\textbf{Display name} & \textbf{Recorded model identifier} & \textbf{Access path/date} & \textbf{Status/notes} \\
\midrule
Claude Sonnet 4 & \path{claude-sonnet-4-20250514} & Anthropic/Claude Code, 2026-03-15 & Complete; no explicit thinking parameter recorded \\
Claude Opus 4.6 & \path{claude-opus-4-6} & Claude CLI 2.1.76, 2026-03-15 & Complete; default CLI thinking behavior \\
Claude Sonnet 4.6 & \path{claude-sonnet-4-6} & Claude CLI 2.1.76, 2026-03-15 & Complete; default CLI thinking behavior \\
GPT-4.1 & \path{openai/gpt-4.1} & OpenRouter, 2026-03-15 & Complete; router alias, snapshot mapping unverified \\
GPT-5.5 & \path{gpt-5.5} & Codex CLI 0.128.0, 2026-05-04 & Complete on 105 observable instances; medium reasoning; JSON output schema; no temperature flag \\
GPT-5.4 & \path{gpt-5.4} & Codex CLI 0.128.0, 2026-05-04 & Complete on 105 observable instances; medium reasoning; JSON output schema; no temperature flag \\
GPT-5.2 & \path{gpt-5.2} & Codex CLI 0.128.0, 2026-05-04 & Complete on 105 observable instances; medium reasoning; JSON output schema; no temperature flag \\
DeepSeek R1 & \path{deepseek-ai/DeepSeek-R1-0528} & DeepInfra, 2026-03-17 & Complete; rerun after earlier gateway failures \\
DeepSeek V3.2 & \path{deepseek-ai/DeepSeek-V3.2} & DeepInfra, 2026-03-15 & Complete \\
Qwen 3 235B & \path{Qwen/Qwen3-235B-A22B-Instruct-2507} & DeepInfra, 2026-03-16 & Complete \\
Llama 4 Maverick & \path{meta-llama/Llama-4-Maverick-17B-128E-Instruct-FP8} & DeepInfra, 2026-03-16 & Complete \\
Llama 4 Scout & \path{meta-llama/Llama-4-Scout-17B-16E-Instruct} & DeepInfra, 2026-03-16 & Complete \\
Gemma 3 27B & \path{google/gemma-3-27b-it} & DeepInfra, 2026-03-16 & Complete \\
Gemini 2.5 Pro & \path{google/gemini-2.5-pro-preview-03-25} & OpenRouter, 2026-03-15 & Incomplete; 40/108 gateway failures; excluded from main table \\
MiniMax M2.5 & \path{MiniMax-M2.5} & MiniMax/OpenAI-compatible API, 2026-03-15 & Complete; many heuristic JSON parses \\
\bottomrule
\end{tabularx}
\end{table}

\section{Error Taxonomy Details}
\label{app:taxonomy}

\Cref{tab:taxonomy_full} provides the complete error taxonomy with examples and the specific line items targeted for each error subtype.

\begin{table}[ht]
\centering
\small
\caption{Complete error taxonomy with targeted line items and example perturbations.}
\label{tab:taxonomy_full}
\begin{tabular}{p{0.08\textwidth}p{0.28\textwidth}p{0.28\textwidth}p{0.25\textwidth}}
\toprule
\textbf{Code} & \textbf{Target} & \textbf{Check} & \textbf{Example} \\
\midrule
AE-Row & BS: component of total assets & Components sum to subtotal & Modify inventory so current assets items $\neq$ total current assets \\
AE-Col & BS: subtotal & Subtotals sum to total & Modify total current assets so subtotals $\neq$ total assets \\
CL-NI/RE & IS: net income & IS net income = $\Delta$ BS retained earnings & Perturb IS net income; RE change now disagrees \\
CL-NI/CFS & IS: net income & IS net income = CFS operations start & Perturb IS net income; CFS start now disagrees \\
CL-Cash & CFS: ending cash & CFS ending cash = BS cash & Perturb CFS ending cash; BS cash now disagrees \\
CL-D\&A & CFS: depreciation & CFS D\&A $\approx$ IS D\&A & Defined but excluded from current release due to inconsistent XBRL reporting \\
YoY-Open & BS: prior year balance & Prior ending = current opening & Modify comparative-year figure \\
YoY-Chg & Any: stated change & $v_t - v_{t-1}$ = stated change & Modify one period value \\
MR-Minor & Any visible numeric field & Magnitude perturbation $<1\%$ & Perturb one field by 0.5\% \\
MR-Mod & Any visible numeric field & Magnitude perturbation 1--5\% & Perturb one field by 2\% \\
MR-Sig & Any visible numeric field & Magnitude perturbation 5--20\% & Perturb one field by 10\% \\
MR-Ext & Any visible numeric field & Magnitude perturbation $>20\%$ & Perturb one field by 25\% \\
\bottomrule
\end{tabular}
\end{table}

\end{document}